\documentclass[twoside, onecolumn]{article}
\usepackage{graphicx} % Required for inserting images
\usepackage{tabularx}
\usepackage{adjustbox}
\usepackage{hyperref}
\usepackage{xr}
\usepackage{enumitem}
\usepackage{amsmath,amssymb}
\usepackage{ragged2e}
\usepackage{biblatex}
\usepackage[top=3cm,bottom=3cm,left=2cm,right=2cm,asymmetric]{geometry}

\usepackage{makecell}

\usepackage[dvipsnames]{xcolor}
\usepackage[rightcaption]{sidecap}
\usepackage{booktabs}
\usepackage{multirow}
\usepackage{soul}
\usepackage{float}
\usepackage{hyperref}
\usepackage{acro}
\DeclareAcronym{nPA}{
    short = nPA, 
    long = neuromorphic-Perception-to-Action
    }

\DeclareAcronym{sPol}{
  short = strike policy,
  long  = strike policy
}

\DeclareAcronym{sP}{
short = sP,
long = spiking-based pinball
}

\usepackage[absolute,overlay]{textpos}
\usepackage[most]{tcolorbox}

\usepackage{xr}

\title{Can Spiking Neural Networks play pinball? A neuromorphic motion detector for target tracking}
\author{
\textsc{Mazdak~Fatahi}\thanks{Corresponding Authors; Co-first authors on this work.}$\text{  }^1$,
\textsc{Šárka Pryjmaková}$^{*2}$,
\textsc{Pierre~Boulet}$^1$,
\textsc{Giulia D'Angelo}$^2$
\\[1ex]
\normalsize $^1$Univ. Lille, CNRS, Centrale Lille, UMR 9189 CRIStAL, F-59000 Lille, France \\
\normalsize $^2$Department of Cybernetics, Faculty of Electrical Engineering, Czech Technical University in Prague \\
\normalsize \href{mailto:Mazdak.Fatahi@univ-lille.fr}{mazdak.fatahi@univ-lille.fr},
\href{mailto:sarka.pryjmakova@cvut.cz}{sarka.pryjmakova@cvut.cz},
\href{mailto:Pierre.Boulet@univ-lille.fr}{Pierre.Boulet@univ-lille.fr},
\href{mailto:giulia.dangelo@fel.cvut.cz}{giulia.dangelo@fel.cvut.cz}
}
\date{}

\begin{document}
\maketitle

\begin{abstract}
Biological visual systems achieve continuous, low-latency motion perception by processing sparse, asynchronous spiking signals, enabling real-time tracking under strict energy constraints. Event-based cameras, inspired by the mammalian retina, replicate this efficiency by capturing only local brightness changes as asynchronous events, offering a natural substrate for spiking neural networks (SNNs) to parallelise computation and adapt to fast-changing scenes. Pinball provides a controlled yet dynamic testbed, requiring precise motion estimation and fast reaction to a small, rapidly moving target. This work presents a fully spiking, real-time perception-to-action pipeline for closed-loop pinball gameplay. A dynamic vision sensor observes a small, fast-moving ball, and a network of spiking Time-Difference Encoders deployed on the SpiNNaker neuromorphic platform jointly estimates its position, speed, and direction. The system is characterised across receptive field size, accumulation window, and angular tuning width for real-time operation, and benchmarked in closed loop against human players across two flipper regimes of increasing physical realism. The system achieves a hit rate of $56.1\%$, nearly double the human average, while reacting within $21.7\,\mathrm{ms}$ (with $5\,\mathrm{ms}$ network latency) and consuming an estimated $148\,\mu\mathrm{W}$ of dynamic power using fewer than $25\,$k neurons. In the benchmark comparison, it ranks among the fastest and most energy-efficient event-based closed-loop neuromorphic-perception-to-action demonstrators. Under more physically realistic flipper dynamics, the system is able to reproduce the full spectrum of human play styles, from cautious to aggressive, by tuning a single interpretable policy parameter, with no change to the perception pipeline. A physical real-time demonstrator, in which a real ball is tracked and real flippers are actuated in closed loop, confirms that the same perception-to-action principle operates beyond simulation. Its fully spiking, learning-free design ensures reliable, low-latency motion estimation, making it a compact and energy-efficient example of closed-loop, real-time neuromorphic perception-to-action systems.
\end{abstract}

\begin{figure}[t]
  \centering
  \includegraphics[width=1\textwidth]{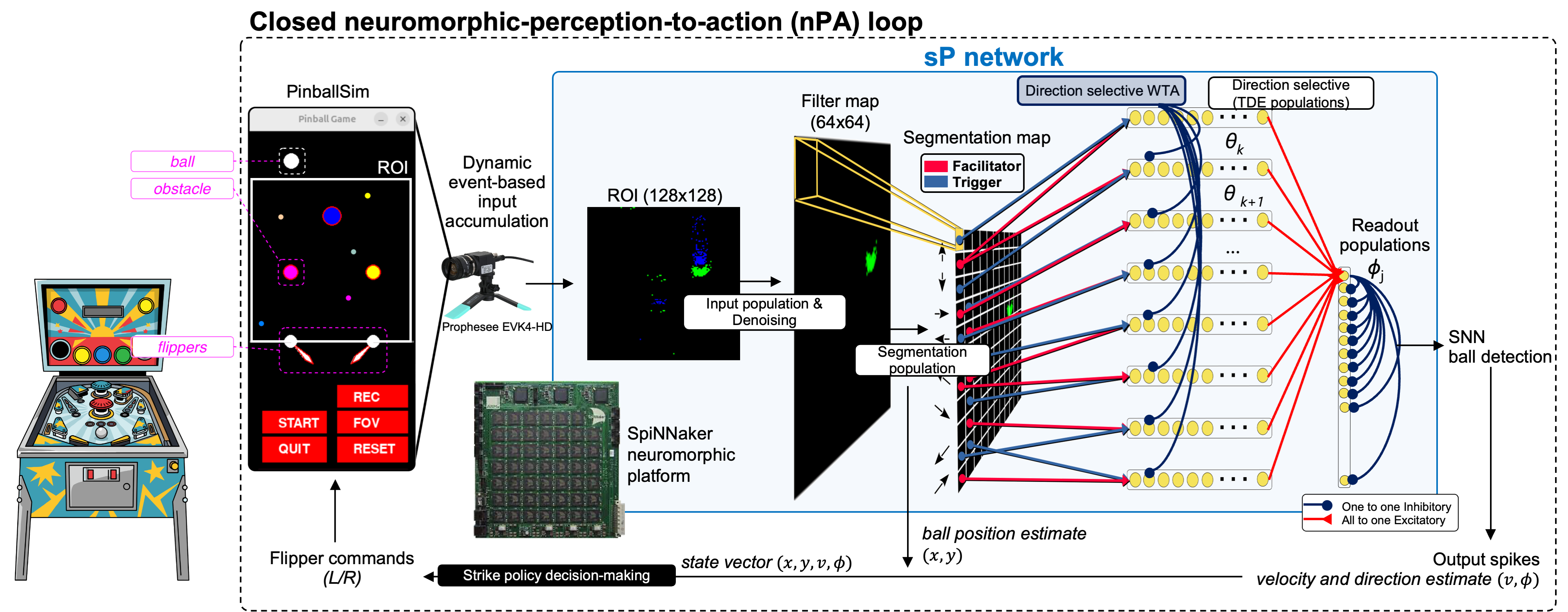}
  \caption{The closed neuromorphic-perception-to-action (\acs{nPA}) loop for real-time pinball gameplay. The pinball simulator (PinballSim) runs on the host PC and is displayed on a monitor, where a Prophesee EVK4-HD event camera records a $128\times128$ region of interest around the flippers and drain. The resulting event stream is processed by the spiking-based pinball (\acs{sP}) network on SpiNNaker: after input accumulation and denoising, a segmentation stage groups the filter map into receptive fields feeding eight direction-selective (TDE) populations, which encode local motion; these project to a readout population that estimates motion direction and speed, while ball position is read from the filter map. The resulting state vector $(x, y, v, \phi)$ is passed to the host-side strike policy, which issues the left/right flipper commands that close the loop.}
  \label{fig:closed-loop}
\end{figure}

\section{Introduction}
%% Problem statement and traditional pipeline in motion estimation and object tracking

The detection of rapid motion and accurate estimation of object velocity constitute fundamental prerequisites for downstream tasks such as object tracking, activity recognition, and collision avoidance in dynamic scene understanding. Consequently, a large and growing body of research in computer vision has leveraged advances in machine learning and deep learning to address these challenges, achieving high accuracy across edge and embedded applications ranging from autonomous driving and industrial automation to elderly fall detection and surveillance systems~\cite{sultani2018real, gu2021survey, wang2020elderly}.
Despite their impressive accuracy, conventional approaches rely on traditional frame-based cameras that process all pixels in every frame, including redundant regions that remain unchanged. For edge and embedded systems with limited memory, power, and compute resources, this leads to unnecessary energy consumption~\cite{somvanshi2025tiny, essahraui2026comprehensive}. Moreover, frame-based sensors inherently face a latency–bandwidth trade-off and often struggle under fast motion, low light, or high dynamic range conditions, where motion blur, poor contrast, and limited dynamic range degrade detection reliability~\cite{zhang2022tracking,zhou2024deep, chen2020event, gallego2020event, maqueda2018event}.

%% event-based vision
In contrast, biological visual systems achieve continuous motion processing with remarkable efficiency. Inspired by biological processes, neuromorphic vision sensors have emerged as promising alternatives to conventional frame-based cameras~\cite{gallego2020event, lichesteiner2008128}. Unlike traditional sensors that capture frames at fixed rates, Dynamic Vision Sensors (DVS) encode pixel-level brightness changes as asynchronous events with microsecond-level temporal resolution, high dynamic range, and minimal latency~\cite{lichesteiner2008128, delbruck2008frame}.

%%  applications and drawbacks of event-based + conventional optical flow estimation on cpu/gpu
Recent work on event-based motion estimation focused on optical flow, typically building dense flow fields rather than local, cell-specific detectors. Benosman et al.~\cite{clady2015asynchronous, benosman2013event} made significant contributions in this direction, introducing event-driven flow formulations and asynchronous feature methods.
More recently, Bardow et al.~\cite{bardow2016simultaneous} proposed the first simultaneous intensity and flow estimation framework, and later Zhu et al.~\cite{zhu2018ev} presented EV-FlowNet, a deep-learning model trained end-to-end on events. The development of large-scale event datasets such as MVSEC~\cite{zhu2018multivehicle} and DSEC~\cite{gehrig2021dsec} further accelerated progress in motion estimation and evaluation. Other recent work from Gao et al.~\cite{gao2022contrast} focuses on contrast-maximisation–based optimisation methods for the estimation of optical flow, evaluating the method in several public object tracking scenarios. While these methods achieve rich visual motion representations, they typically require offline training, large memory footprints, or computationally expensive processing, failing to exploit the inherent sparsity and efficiency of biologically plausible spiking-based neuromorphic systems.

%% biological inspiration to resolve the issues with cpu/gpu event-driven models in motion detection
Meanwhile, biological vision systems exhibit efficient direction selectivity much earlier in the processing pipeline, as early as retinal ganglion cells~\cite{wang2023type}, followed by specialised motion analysis in cortical areas MT/V5 and MST~\cite{wurtz2008neuronal, born2005structure}. These neurons exploit sparse, asynchronous spiking signals to encode motion direction, velocity, and object trajectories with remarkable computational efficiency, processing only changes in the visual scene via temporal contrast rather than dense pixel grids~\cite{felleman1991distributed, borst2011seeing}. This biologically plausible approach enables low-latency, low-power parallel computation critical for real-time survival tasks like pursuit and tracking~\cite{newsome1988selective}, unlike the continuous, memory-intensive representations of event-based deep networks.

%% Bio-inspired abstract model for efficiently processing the event-based information ( in contrast to processing events with conventional cpu-based algorithms ): EMD, energy-based model and EMD in fly and Drosophila
Motion perception in artificial visual systems has been strongly inspired by biological models of early vision~\cite{movshon1985analysis, borst2011seeing, wei2018neural}. Drawing on direction selectivity in biological systems, early foundational work, such as the Hassenstein–Reichardt detector~\cite{hassenstein1956systemtheoretische} and subsequent extensions by Borst and Egelhaaf~\cite{borst1987temporal}, established the concept of correlation-based Elementary Motion Detectors (EMDs) as the simplest mechanism for computing local motion through spatiotemporal correlation. In parallel, the spatiotemporal energy model from Adelson and Bergen~\cite{adelson1985} offered a different computational basis for motion direction selectivity, shaping motion direction detection by measuring \textit{energy} from filters tuned to specific spatiotemporal patterns in visual inputs. Later, Borst and Euler~\cite{borst2002neural} refined the classic Hassenstein-Reichardt correlator~\cite{hassenstein1956systemtheoretische} into a canonical model for correlator-based motion perception using elementary motion detectors (EMDs) in fly visual circuits. Their work~\cite{borst2002neural} incorporated ON/OFF signal segregation to explain flicker suppression and contrast invariance.
Further studies in Drosophila have later uncovered specialised small-target–selective and object-detecting neurons, inspiring artificial mechanisms for attention, feature selection, and object tracking in neuromorphic systems~\cite{yang2018elementary}.

%% EMD+input frame+CPU: inefficient, EMD+DVS+Neuromorphic processing: perfect
While these bio-inspired models capture the efficiency of biological motion detection, most conventional implementations process dense frame-based inputs rather than sparse asynchronous events from neuromorphic sensors. They are also typically executed on traditional digital computing architectures, limiting the benefits of event-driven processing.
% neuromorphic hardware
Neuromorphic hardware~\cite{douglas1995neuromorphic}, including both analog~\cite{pehle2022brainscales,moradi2017scalable,qiao2015reconfigurable} and digital platforms~\cite{Furber2014,davies2018loihi,sga2020neuromorphic}, addresses these limitations by integrating memory and computation. This architecture avoids the von Neumann bottleneck and supports highly parallel, event-driven processing. On such platforms, Spiking Neural Networks (SNNs) efficiently process asynchronous event streams~\cite{DiehlS2015SpikingDeepNetworks,7838165,gehrig2020eventbasedangularvelocityregression,SpikeMS}, enabling low-latency and energy-efficient real-time operation~\cite{fatahi2024event,d2020event}. Their event-driven operation reduces data movement and power consumption. Platforms such as SpiNNaker~\cite{furber2014spinnaker,hopkins2018spiking}, Loihi~\cite{davies2018loihi}, and FPGA-based systems~\cite{nasirneuromorphic,saulquin2025modnef} enable efficient real-time motion processing with event-based sensors~\cite{fatahi2024event,d2020event,glover2019atis}.

%% neuromorphic developments: EMD
Early neuromorphic motion-processing systems focused on local motion estimation using variants of the Reichardt correlator, enabling direction selectivity and velocity estimation from event streams~\cite{giulioni2016event,benosman2013event}. These approaches were primarily applied to robotics tasks such as navigation and obstacle avoidance~\cite{giulioni2016event,lagorce2016hots}.
%% sEMD and TDE and neuromorphic applications
More recent work has extended neuromorphic motion detection beyond classical Reichardt correlators. Milde et al.~\cite{milde2018spiking} introduced the spiking Elementary Motion Detector (sEMD), providing a fully spike-based implementation of delay-correlator motion sensing. This principle was later generalised to multi-point Temporal Difference Encoder (TDE) frameworks~\cite{yedutenko2025tde}. Gutierrez-Galan et al.~\cite{gutierrez2021event} proposed a digital TDE model and demonstrated efficient FPGA implementations for several localisation and navigation tasks. Other studies explored biologically inspired extensions, including eccentricity-dependent motion detection~\cite{d2020event} and the emergence of direction selectivity through adaptive synaptic delays~\cite{grimaldi2023}.

%% limitations of these works and our contribution
Previous EMD/TDE studies typically evaluate direction selectivity using large moving bars, drifting gratings, or wide-field stimuli, which activate broad regions of the sensor. Consequently, they mainly assess coarse left-right or up-down selectivity rather than precise spatial localisation. Obstacle-gap detection in cluttered scenes is a case in point, where many neurons are activated simultaneously, and individual receptive fields (RFs) are rarely probed in isolation~\cite{giulioni2016event,d2020event,fschoepe2024finding}. Studies using small moving targets~\cite{delbruck2013robotic} fare little better, as evaluation still centers on global response consistency or preferred-direction estimation, not on precise tracking of small objects across space.
Taking inspiration from previous implementations~\cite{fschoepe2024finding,d2020event,milde2018spiking}, this work evaluates motion selectivity under local activation. Only a small subset of RFs is active as a target moves across the field of view, providing a stricter test of local motion encoding than the wide-field stimuli discussed above. 
The pinball game represents a controlled toy problem for this class of tasks: real-world sensorimotor scenarios, such as tracking small, fast-moving objects for collision avoidance or interactive robotic control, similarly demand precise local motion and trajectory estimation from sparse activation, rather than coarse directional classification. 
Such problems have been pursued more broadly by the event-based community, including robotic goalkeeping~\cite{robotgoalie_delbruck,robotgoalie_cheng} and high-speed ball catching~\cite{wang_catch}, closed-loop play of the Atari Pong game~\cite{pong_dreaming_blakowski,Rizzo_2025} and its physical, higher-dimensional counterpart, air hockey~\cite{Gava_PUCK,Romero_hockey,air_hockey_ambrosini}, as well as balancing tasks such as pencil balancing~\cite{Conradt_pencil}.
Most exploit event-based sensing, and several additionally process the events with spiking neural networks~\cite{robotgoalie_cheng,ziegler_tennis,pong_dreaming_blakowski,Rizzo_2025,Romero_hockey,air_hockey_ambrosini}; however, only a few combine event-camera input, spiking-based processing, and deployment on neuromorphic hardware in a closed loop~\cite{robotgoalie_cheng,Romero_hockey}.
The proposed spiking-based pinball (\acs{sP}) network falls into this last category, fully neuromorphic, spike-driven, and running on-chip in low-power closed-loop control, while additionally targeting fine-grained motion estimation from local receptive fields activation.

%% Follow up and our toy problem
Building on our prior work on event-based motion processing~\cite{fatahi2024event,d2020event}, the proposed work extends SNN-based motion processing to a controlled yet dynamic \textit{pinball} environment, relying on spike-based EMD frameworks in which TDE units form the basic computation block~\cite{milde2018spiking,fschoepe2024finding}. A standard pinball machine provides a highly dynamic and noisy visual environment. Fast motion, flashing lights, and mechanical elements generate dense event streams. These factors make reliable tracking difficult under strict latency and energy constraints. To the best of the authors' knowledge, this is the first TDE-based SNN that (i) extracts fine-grained angular motion from sparse activations, (ii) estimates velocity and direction jointly in real time, and (iii) demonstrates closed-loop control in a dynamic pinball system. The main contributions are summarised as follows:

\begin{itemize}

\item A configurable, physics-based pinball simulator and event-acquisition pipeline, comprising a minimal variant with ground-truth position and velocity for controlled system characterisation, an interactive variant with two flipper regimes for closed-loop evaluation, and an adaptive event-count packaging (AECP) strategy that mitigates motion-induced blur at high speed while avoiding excessive latency at low speed.

\item A neuromorphic motion-detection architecture, in which local Time Difference Encoders (TDEs)~\cite{milde2018spiking, clark2011defining, mauss2015neural, Chicca} interact through eight directional pathways and project to a continuous 36-neuron angular readout layer for fine-grained motion estimation, implemented entirely through structured spiking computation with no learned or trained parameters, and fully deployed on the SpiNNaker neuromorphic platform for real-time, on-board motion decoding from raw DVS events.

\item Benchmarking of the resulting latency and power figures against prior event-based closed-loop demonstrators (Table~\ref{tab:toy_examples}), showing the system to be competitive with, and in several cases faster or more energy-efficient than, existing neuromorphic sensorimotor systems.

\item Quantitative benchmarking of the closed-loop system against human players, achieving a hit rate more than double the human average in the binary-gate regime, and showing, in the solid-flipper regime, that a single decision-policy parameter reproduces the full spectrum of human play styles, from cautious to aggressive, without modifying the perception pipeline.

\item Validation of the approach beyond simulation with a physical demonstrator, in which a real metal ball is tracked by the DVS, processed by the SNN, and used to actuate physical flippers in closed loop, confirming that the same perception-to-action principle operates on real-world hardware.
\end{itemize}

\section{Methods}\label{ch:methods}

%% Entire of the pipeline is shown in Figure-Full_Pipeline_full_tracker: 
%% Stages
%% I. Perception, 
%% II. decision making and control from position estimate -> action

%% STAGES - architecture steps:
%% Stage_1:simulator -> Stage_2:camera and dynamic input -> Stage_3:SNN (gaussian-denoising, segmentation and RFs, direction-selection, read-out:andgles and wta) -> Stage_4:data colecting and deision making

%% Intro to the subsections (stages of the pipeline)
The proposed neuromorphic-perception-to-action loop (\acs{nPA}), illustrated in Figure~\ref{fig:closed-loop}, operates as a closed-loop architecture that couples event-based perception to pinball game control. It comprises two functional stages. In the first, fully spiking-based perception stage, a DVS camera captures the ball in motion across the simulated pinball playfield, using a custom, open-source pinball simulator (PinballSim) developed for this work\footnote{Further information about the Data and Code Availability in Section~\ref{ch:code}}, and the resulting events are streamed in real time to a spiking neural model (\acs{sP} network) running on the SpiNNaker neuromorphic platform. The \acs{sP} network processes the raw event stream to derive an estimate of the ball's state vector $(x, y, v, \phi)$, where $(x, y)$ denotes the detected ball position, $v$ is the estimated speed, and $\phi$ is the motion direction. The resulting state estimate is then passed to the second, decision-making stage, where \textit{\acs{sPol}} maps it to a control command for flipper actuation. The model processes these raw event streams to extract motion features, estimate the ball’s velocity vector, and generate control decisions that are translated into commands for actuating the flippers.
trackers

\subsection{Input acquisition}

The perception stage begins with generating and capturing the visual stimulus that drives the \acs{sP} network. This is achieved in two steps: a custom pinball simulator renders the game and provides ground-truth position and velocity for evaluation (Section~\ref{ch:sim_recordings}), while a DVS camera observes this rendered scene and converts it into the accumulated event packages that form the network's raw input (Section~\ref{ch:dyn_event_acc}).

%% 1st Stage: Game simulators
\subsubsection{The pinball game simulator: PinballSim}\label{ch:sim_recordings}

\begin{figure}[htbp]
  \centering
  \includegraphics[width=0.75\textwidth]{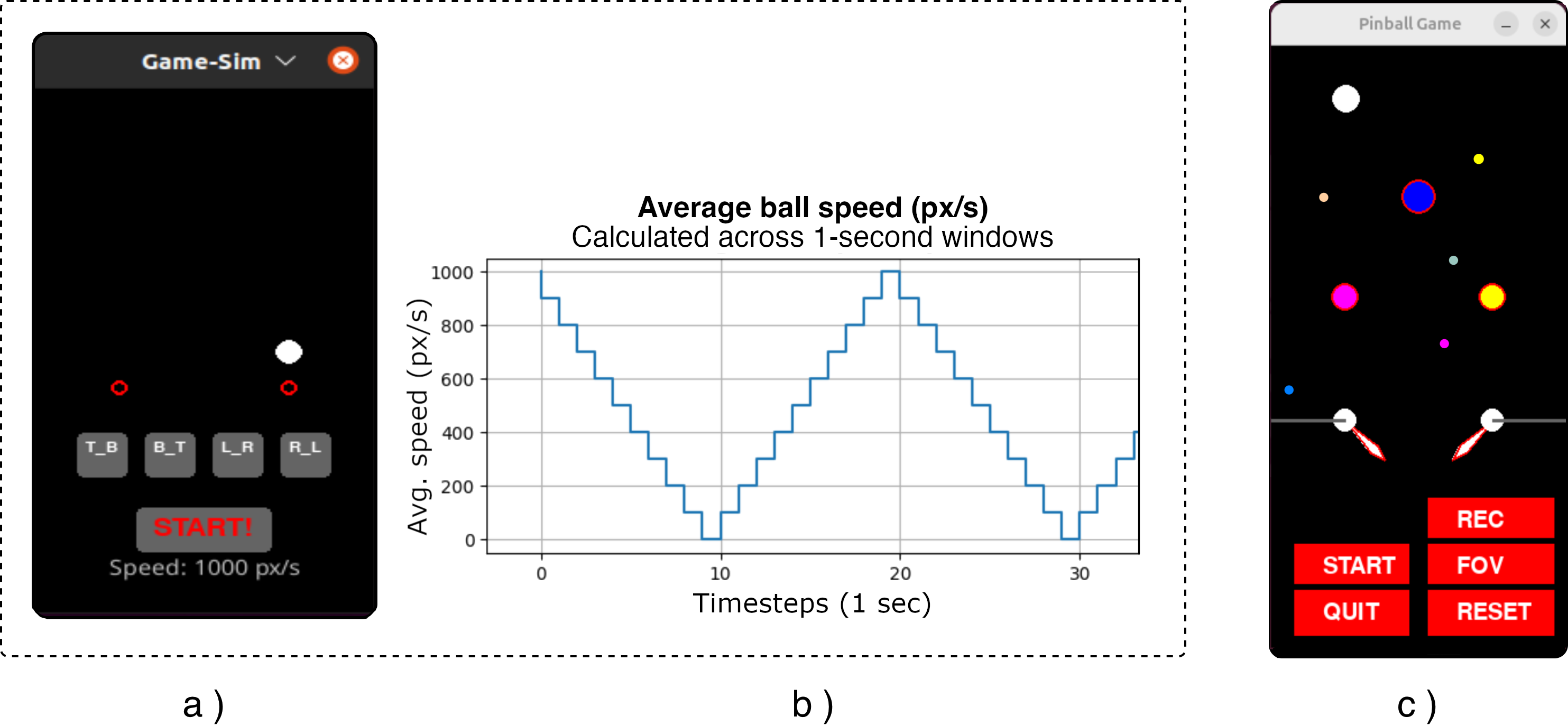}
  \caption{Pinball simulator variations. a) Minimal game simulator, in which the ball follows predefined trajectories used to characterise the \ac{sP} network. b) Ball-speed profile in the minimal simulator: constant speeds from 0 to 1000~px/s, stepped in 100~px/s increments and held for 1~s each, repeated over successive cycles. c) Interactive game simulator, a fully playable environment used for closed-loop evaluation.}
  \label{fig:simulators}
\end{figure}

A DVS camera produces the event-based input by observing a custom, physics-based pinball simulator (PinballSim) rendered on a monitor. It provides a reproducible, configurable environment for studying motion perception and closed-loop control. Two versions of PinballSim are implemented: a simplified \textit{minimal game simulator}, used to calibrate the acquisition pipeline and model parameters, and a complete \textit{interactive pinball game simulator}, used for closed-loop evaluation (see Figure~\ref{fig:simulators}). The interactive version additionally implements two flipper regimes, binary-gate and solid flipper. PinballSim is implemented using the \texttt{Pygame} library~\cite{pygame2024}\footnote{https://github.com/pygame/pygame} and records the ball's position and velocity to generate ground truth (the link to the code can be found in Section~\ref{ch:code}).

%% Controlled simulator: 8 main directions
The \textit{minimal game simulator} version of the game was designed for system characterisation purposes and preserves the essential motion dynamics of pinball while deliberately removing unnecessary visual and mechanical complexity, such as reflections and moving objects. In this controlled configuration, the ball moves exclusively along predefined trajectories. Linear motion is generated along eight discrete directions (0°, 45°, 90°, …, 315°) 
to systematically evaluate the sensitivity and selectivity of eight direction-selective populations within the \ac{sP} network (Section~\ref{ch:Neuromorphic Processing: Network Architecture}), as well as their ability to estimate the speed and direction of motion. In addition, a circular trajectory of the ball, executed in both clockwise and counter-clockwise directions, is used to assess the behaviour of \acs{sP}'s angular readout population (For further details, see Section~\ref{ch:Neuromorphic Processing: Network Architecture} and Figure~\ref{fig:closed-loop}), validating its capacity to produce a continuous angle representation over a full 360° motion cycle.

%% Full Simulator
The \textit{interactive game simulator} (Figure~\ref{fig:simulators}C) extends the minimal version into a full playable environment, restoring the visual and mechanical complexity omitted during the characterisation. Gravity, continuous friction, and energy loss on wall and flipper contacts all act on the ball, so its speed evolves dynamically throughout play. Two paddle-like flippers at the base are driven by the \acs{sPol} (See details in Section~\ref{ch:strikePolicy}) to keep the ball in play. The interactive simulator supports two flipper regimes of increasing physical realism. In the \textit{binary-gate} regime, a flipper acts only at the instant it is triggered, and has no effect on the ball otherwise, isolating the perception-action chain from post-contact physics. In the \textit{solid flipper} regime, the flippers are persistent colliders that occupy the playfield at all times and deflect the ball on contact, whether or not actively triggered, thereby reproducing the contact dynamics a human player intuits. These two regimes define the two closed-loop experiments of Section~\ref{ch:closed-loop}.

\subsubsection{Dynamic event-based input accumulation}\label{ch:dyn_event_acc}
%% 2nd Stage: Recording configuration and input pre-processing

The stream of events is acquired using a Prophesee EVK4-HD event-based camera~\cite{prophesee_evk4} ($1280\times720$ px), positioned $\sim$75~cm in front of an HP P24q G4 HD monitor displaying the pinball environment. Rather than the full playfield, the camera is configured to record only a $128\times128$ px region of interest (ROI) around the flippers and drain, marked by the white square in Figure~\ref{fig:closed-loop} (introduced in Section~\ref{ch:methods}). This choice reflects the structure of the task; the decisive moment occurs as the ball approaches the lower region, so tracking the ball across the entire playfield is unnecessary. The ROI nonetheless spans a substantial part of the playfield, providing sufficient context for tracking while reducing the input size, and with it the number of required input neurons and synaptic connections, resulting in a smaller, more energy-efficient spiking network suited for resource-constrained neuromorphic hardware.

\begin{figure}[h!]
    \centering
      \includegraphics[width=0.7\textwidth]{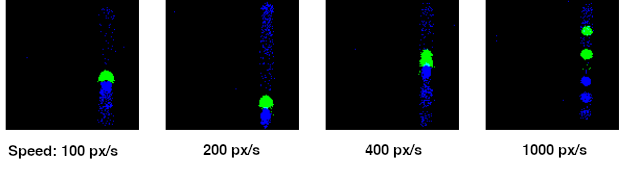}
  \caption{Effect of temporal accumulation for a fixed accumulation window on pseudo-frames at increasing ball speeds. Pseudo-frames are formed by accumulating positive events over $\Delta T = 20$~ms. Green dots indicate positive events. At low speeds, events form a compact representation of the ball, while at higher speeds, the pseudo-frame exhibits elongated or multiple clusters, illustrating motion-induced temporal overlap and smearing.}
  \label{fig:pseudo_frame}
\end{figure}

Each event $(x,y)$ within the ROI drives the Input Population, which generates the spiking activity of the \acs{sP} network, and is streamed to the SpiNNaker board over Ethernet as a real-time, low-latency input. These events are accumulated into packages for the model, either over a constant time interval ($\Delta T$) or over a fixed number of events.
%% CTI: Constant Time Interval and the issues with high speed
A constant time interval (CTI) accumulation window ($\Delta T$) is initially adopted to forward events to the model. At higher ball speeds, however, this approach introduces temporal overlap: within the same $\Delta T$ interval, events may correspond to multiple spatial locations of the ball, yielding a blurred trajectory rather than a discrete snapshot.
As shown in Figure~\ref{fig:pseudo_frame}, at 1000~px/s and 60~Hz refresh rate, a pseudo-frame shows multiple event clusters corresponding to different ball positions, representing its trajectory rather than an instantaneous state (without the refresh-rate constraint, the motion would appear continuous, forming a single uninterrupted streak instead of discrete clusters).
Since TDE cells rely on precise temporal ordering across adjacent receptive fields, this behaviour degrades the spiking motion estimation network, where multiple ball positions within a single pseudo-frame disrupt the facilitation-trigger sequence. This effect is most pronounced when the receptive field is smaller than the displacement over one $\Delta T$, allowing the ball to cross several segments between updates. Reducing $\Delta T$ only partially mitigates this issue; while shorter windows reduce motion overlap, they can produce sparse input for slow motion and may fail to capture sufficient information. Moreover, smaller $\Delta T$ increases the transfer rate to the SpiNNaker board, raising communication overhead and risking Ethernet congestion.

\begin{figure}[htbp!]
  \centering
    \includegraphics[width=0.65\textwidth]{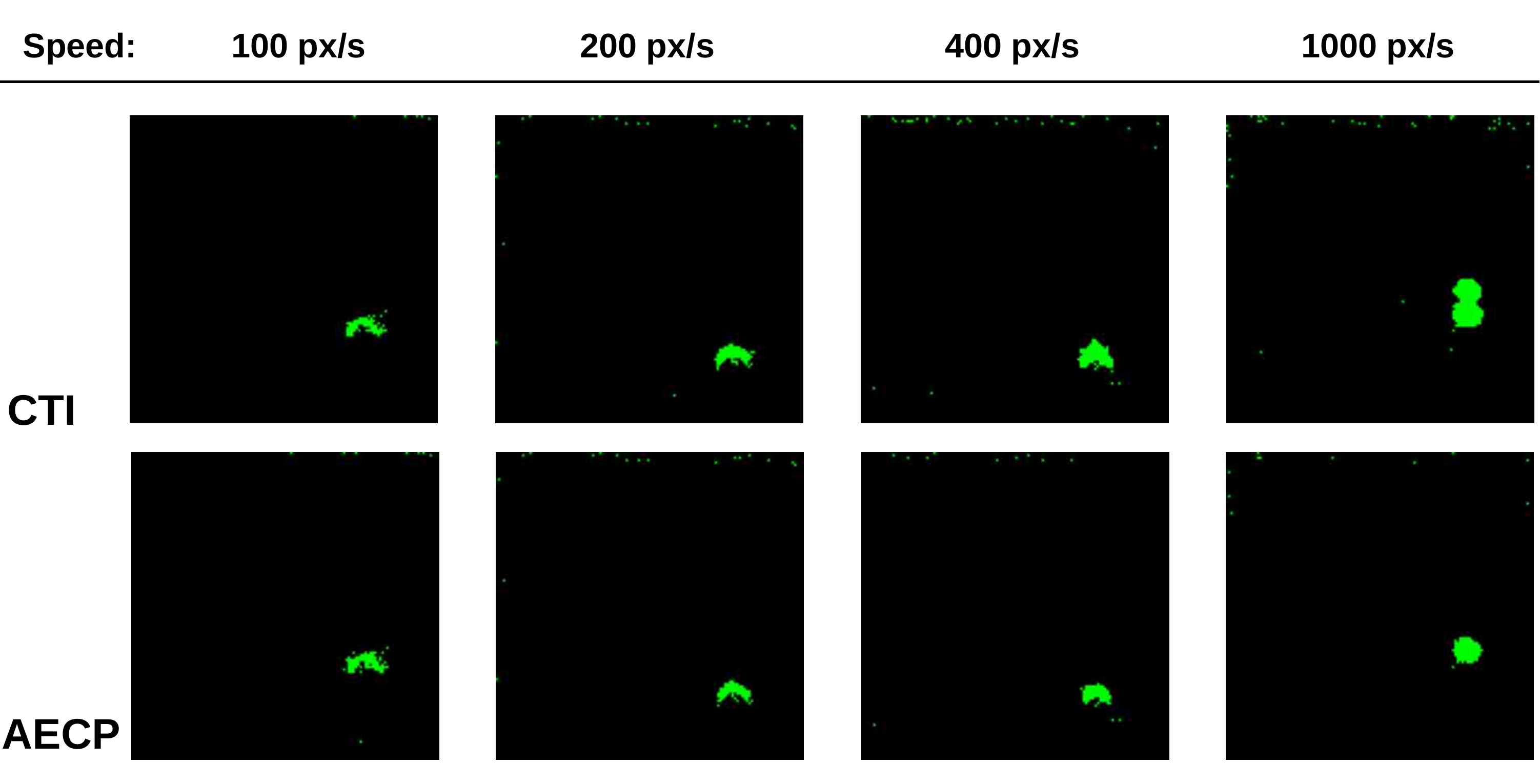}

  \caption{Comparison between Constant Time Interval (CTI) accumulation and the Adaptive Event-Count Framing (AECP) method at different ball velocities (100, 200, 400, and 1000 px/s). Each pseudo-frame displays 200 positive events (not showing the negative events) accumulated within either a fixed temporal window ($\Delta T = 20$ ms, top row) or an adaptive window determined by the AECP strategy (bottom row). In the CTI case, higher speeds cause elongated or duplicated event clusters due to temporal overlap, resulting in motion-induced blur. In contrast, the adaptive approach automatically shortens $\Delta T$ at higher velocities, preserving compact and coherent event representations while maintaining temporal consistency across motion conditions.}
  \label{fig:cti_ada}
\end{figure}

%%% CNE: Constant Number of Events and issues
An alternative collection of events relies on event counting, which generates a package of events after a fixed number of events ($N_{evt}$). This approach adapts to motion speed; however, in real-time use, slow motion can increase latency due to the time required to accumulate enough events. In this work, an adaptive event-count package method (AECP) is proposed; event packages are collected by combining event-count and time-bounded accumulation, using either $N_{evt}$ or a maximum time limit $\Delta T_{max}$. In the proposed Adaptive event-count packaging (AECP) method, events are accumulated using two complementary constraints: a fixed number of events, $N_{\mathrm{evt}}$, and a maximum accumulation time, $\Delta T_{\mathrm{max}}$. Event accumulation stops as soon as either constraint is satisfied, with the first condition reached determining the end of the current package of events. During periods of rapid motion, the event generation rate increases significantly, allowing the threshold $N_{\mathrm{evt}}$ to be reached before $\Delta T_{\mathrm{max}}$. Consequently, the accumulation window is adaptively shortened, resulting in lower latency while preserving sufficient spatial information. Conversely, during periods of low activity, when events are generated more sparsely, $\Delta T_{\mathrm{max}}$ guarantees that a package of events is produced within a bounded time, preventing excessive delays in downstream processing. This hybrid accumulation strategy provides an effective trade-off between low latency and temporally consistent neuromorphic input by adaptively switching between event-driven and time-driven framing according to the scene dynamics (see Figure~\ref{fig:cti_ada}).

\subsection{Spiking-based pinball (sP) network}\label{ch:Neuromorphic Processing: Network Architecture}
%% 3rd stage: SNN model (Input and Denoising populations, Segmentation map and RFs, DS populations, Continuous angle readout)

%% NOTE: neuromorphic closed-loop (ncl) and spiking pipeline (sP) are not the same: ncl include the game simulators and iteraction btw the board and cpu, but sP is only the SNN running on SpiNNaker

The spiking-based pinball (sP) network (Figure~\ref{fig:closed-loop}) transforms the raw event stream (\ref{ch:sim_recordings}) into a fully spiking, real-time estimate of the ball's state, comprising its position $(x,y)$, speed $v$, and direction of motion $\phi$. The \acs{sP} network consists of four successive population layers: Input \& Denoising, Filter \& Segmentation, Direction-Selective, and Readout populations. Each stage is detailed in the following subsections.

The SNN sP model is implemented and developed using the PyNN framework~\cite{davison2009pynn}, and deployed on a SpiNNaker-5 neuromorphic computing platform~\cite{furber2014spinnaker}. The SpiNNaker-5 board is a massively parallel low-power architecture designed specifically for large-scale spiking neural network simulation in real time. Each board contains multiple ARM-based processing cores interconnected through an asynchronous packet-based communication fabric that natively supports spike routing with high reliability and extremely low latency. This architecture enables efficient emulation of biologically inspired neural dynamics and large-scale connectivity patterns while maintaining deterministic timing. These properties are essential for real-time neuromorphic robotics and closed-loop interaction with sensory input streams such as DVS event data.

%% input stage 
At the input stage, positive events are accumulated using the hybrid adaptive event-count package (AECP) approach, with a $\Delta T$ empirically determined during characterisation and a fixed cap of $N_{\mathrm{evt}} = 500$ events per package. This upper bound reflects the ball's size (radius = 15 px), which occupies approximately $1\%$ of the selected region of interest ($256 \times 256$ pixels) of the pinball playfield (see Section~\ref{ch:syst_charact} for parameter selection details). Restricting accumulation to positive (leading-edge) events halves the input event rate, further reducing the computational load of downstream processing without discarding motion-relevant information, since the trailing edge of the ball's silhouette carries largely redundant timing cues. Each event package is first passed through a Gaussian kernel for denoising and smoothing (Section~\ref{ch:Den_pop}), and the smoothed output is then forwarded to the segmentation population, which splits the input space into a grid of non-overlapping receptive fields (RFs), each spanning an $N\times N$ block of neurons. Within each RF, all neurons converge onto a single neuron in the resulting segmentation map, so that the original input is represented at a coarser spatial resolution downstream. The RF size (N) is a configurable parameter whose impact on system performance is investigated in Section~\ref{ch:SegmentationRFs}. Pooling within receptive fields both reduces the computational cost of the downstream network and improves robustness to the discontinuous motion artefacts introduced by the simulator's 60~Hz display.

The segmentation map provides input to eight Direction-Selective (DS) populations via structured synaptic connectivity between neighbouring RF cells (see Section~\ref{ch:DS populations}). Each DS population is implemented as a TDE population and is tuned to one of the eight principal directions available within a $2\times2$ RF neighbourhood, yielding an angular resolution of $45^\circ$. While this resolution is sufficient for coarse motion detection, it is inadequate for accurately estimating the trajectory and velocity vector of the ball. The identity of the winning DS population gives a coarse motion direction, and its mean firing rate gives ball speed. To sharpen this estimate, an additional readout layer comprising 36 neurons was introduced.  All eight DS outputs project onto this angle-readout population (Section~\ref{ch:AngleDetectionPopulation}, Equation~\ref{eq:w_angular}), in which each neuron represents a $10^\circ$ interval. This yields a continuous, fine-grained estimate of motion direction and, consequently, a more accurate velocity vector.

\subsubsection{Input and Denoising populations}\label{ch:Den_pop}
%% 3rd stage-1: SNN-input
Each 128$\times$128 package of events is processed by the input population on the SpiNNaker board and passed through a convolutional layer implemented using excitatory synaptic connections. A 3$\times$3 Gaussian kernel (Equation~\ref{eq:gaussian_2d}) with a stride of $(2, 2)$ and the \textit{same padding} is applied, performing both local denoising and spatial downsampling in a single operation. A 3$\times$3 kernel is the smallest size that still captures a meaningful local neighbourhood for smoothing, making it an efficient choice that avoids the excessive information loss associated with larger kernels.

\begin{equation} \label{eq:gaussian_2d}
G(x, y) = \frac{1}{2\pi\sigma^{2}} 
\exp\!\left( -\frac{x^{2} + y^{2}}{2\sigma^{2}} \right),
\qquad x, y \in \{-1, 0, 1\}
\end{equation}

Here, $x$ and $y$ denote the relative coordinates within the $3\times3$ Gaussian kernel, with values in $\{-1, 0, 1\}$. These coordinates specify the spatial offsets around the centre of the kernel and do not correspond to absolute coordinates in the input population. The 
$\sigma$ value is the standard deviation of the Gaussian kernel, and in our implementation is empirically set to $\sigma = 0.8$ to obtain a filter that is neither too sharp nor too flat. The kernel is normalised to ensure intensity preservation:
\[
\sum_{x=-1}^{1} \sum_{y=-1}^{1} G(x, y) = 1,
\]
guaranteeing that the filter behaves as a properly normalised Gaussian.
The convolution operation is given by:

\begin{equation} \label{eq:conv_2d}    
I'(x, y) = 
\sum_{i=-1}^{1} \sum_{j=-1}^{1} 
G(i, j) \cdot I(x + i,\, y + j),
\end{equation}
where $I(x, y)$ represents the spiking activity of the input neurons (accumulated positive events), $G(i, j)$ denotes the Gaussian kernel weights, and $I'(x, y)$ is the filtered postsynaptic activity. The stride 
of $(2,2)$ ensures that every other pixel is sampled in both spatial dimensions, reducing the input resolution from $128\times128$ to $64\times64$.The Gaussian filter acts as a lightweight denoising mechanism, improving the signal-to-noise ratio before segmentation while simultaneously reducing the spatial resolution of the representation. As a result, the filter layer produces a compact, downsampled filter map while processing only relevant spikes in the system. Subsampling therefore emerges naturally from the convolution operation itself, eliminating the need for a separate preprocessing stage.

\subsubsection{Segmentation population}\label{ch:SegmentationRFs}
%% 3rd stage-2:: SNN- Importance of the size of the Receptive Fields 
%% Explaining issue with refresh rate (motion blur has solved by adaptive accumulation, jumping ball by segmentation)
The pinball simulation is displayed on an HP~P24q~G4 monitor operating at a refresh rate of 60~Hz (Section~\ref{ch:dyn_event_acc}), introducing temporal discretisation into the visual stimulus. Although the simulator provides the ground-truth position and velocity of the ball at every simulation step, the displayed motion is constrained by the monitor's refresh interval (16.67~ms). Consequently, the apparent motion observed by the event camera differs from the continuous ground-truth trajectory. At low ball speeds (e.g., 60~px/s), the displacement between consecutive frames is approximately 1 pixel, resulting in smooth, continuous motion. However, at higher speeds (e.g., 600~px/s), the ball can travel up to 10 pixels between successive screen refreshes, producing discontinuous spatial jumps. Event accumulation cannot reconstruct these missing intermediate positions, since no visual information is presented during the skipped intervals. Such discontinuities are particularly detrimental for spiking motion-detection systems, which estimate motion by exploiting the precise temporal relationships of spikes generated across neighbouring receptive fields. As a result, the temporal sampling imposed by the display becomes a limiting factor for reliable motion estimation at high ball velocities, despite the simulator providing accurate continuous ground-truth dynamics. The available 60~Hz monitor was used nonetheless, as it let the DVS observe a realistic scene under the same conditions as the interactive game and physical demonstrator, rather than bypassing the camera via synthetic events.

In the proposed pipeline, the TDE~\cite{Chicca} encodes the temporal difference between stimuli received by neighbouring RFs (i.e. the sEMD~\cite{milde2018spiking, clark2011defining, mauss2015neural}). Each RF's synapse is connected to a single TDE, making RF size the key factor determining the measurable velocity range. Selecting an appropriate RF size is therefore essential for covering the desired range of ball speeds. To characterise the model, the system exploits the dependence of receptive field size on preferred ball speed. Smaller RFs are more sensitive to slow motion, as they require only a small displacement to trigger successive activations across neighbouring units, whereas larger RFs are more robust to the large displacements produced by fast motion. To characterise the model (Section~\ref{ch:syst_charact}) and determine suitable RF dimensions for the full range of expected ball velocities, controlled experiments are conducted using a minimal simulation setup. The appropriate RF size is identified by grouping neighbouring cells during the segmentation process, selecting the configuration that balances spatial precision with robustness to high-speed motion. Although a fully biologically inspired solution would rely on multiple parallel pipelines operating at different spatial scales, this is not feasible under the strict hardware and power constraints of embedded neuromorphic systems. Instead, a compact, resource-efficient design is adopted, evaluating several receptive field grouping sizes and analysing their impact on performance.

\subsubsection{Direction-selective (DS) populations}\label{ch:DS populations}
%% 3rd stage-3: SNN-Dierction Selectivity

\begin{figure}[t]
    \centering
    \begin{minipage}[t]{0.4\textwidth}
        \includegraphics[width=0.95\textwidth]{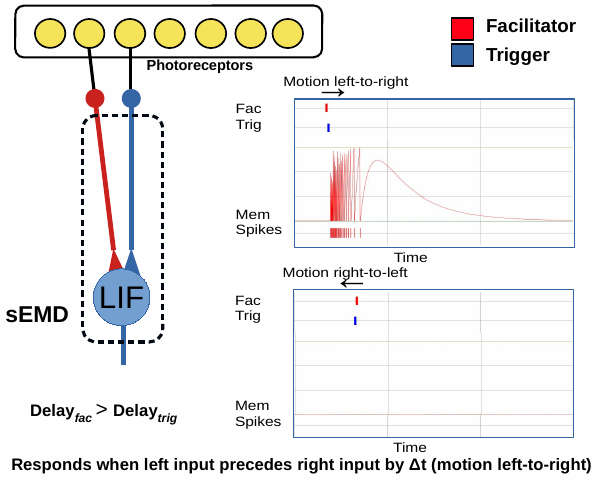}
        \caption{spiking Elementary Motion Detectors (sEMD) characterisation.}
        \label{fig:semd}
    \end{minipage}
    \hfill
    \begin{minipage}[t]{0.55\textwidth}
        \centering
            \includegraphics[width=.85\textwidth]{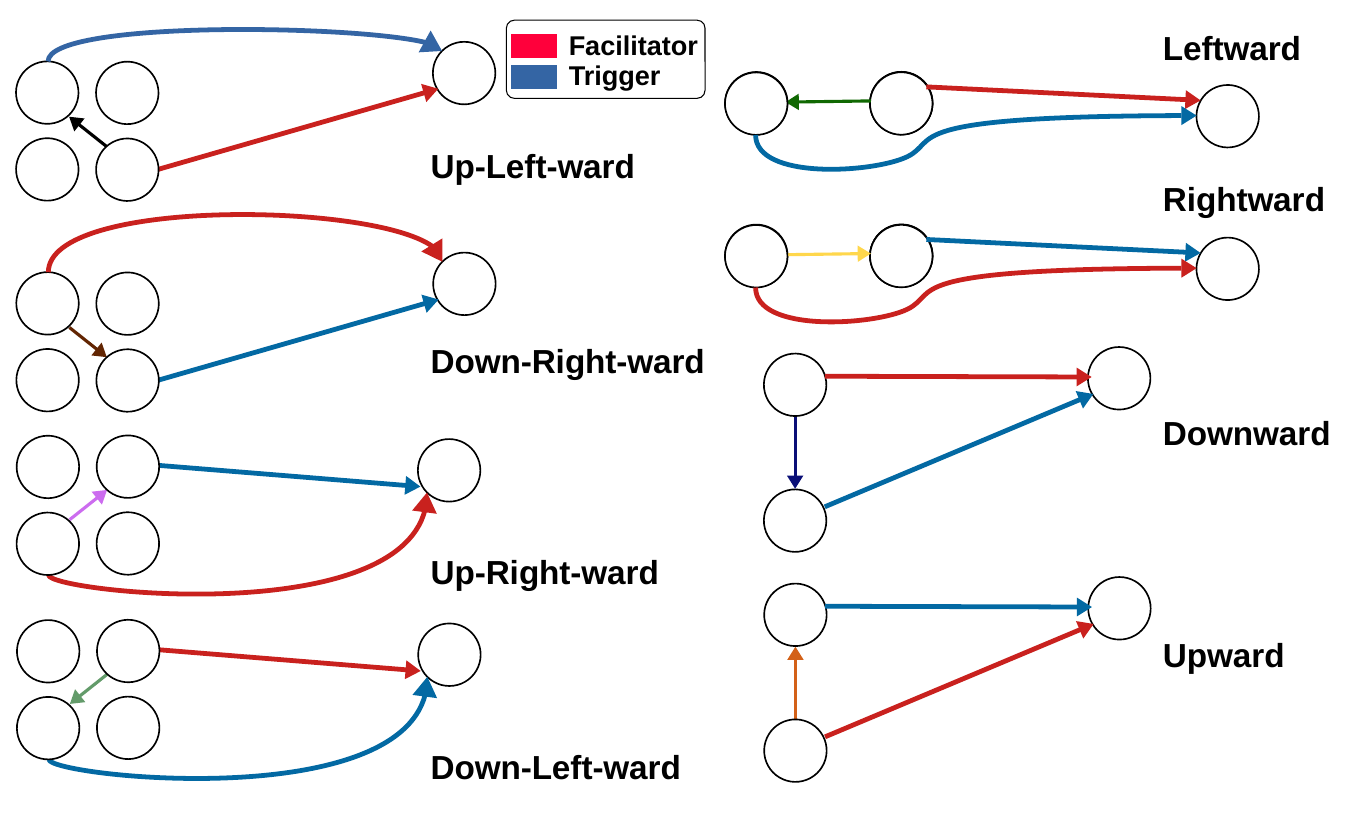}     
        \caption{DS encoding from a $2\times2$ neighbourhood, producing eight motion directions through ordered facilitator–trigger connections.}
        \label{fig:directions}
    \end{minipage}
\end{figure}

The implementation of the direction-selective (DS) populations is inspired by motion-processing mechanisms observed in areas V1 and MT/V5 of the primate visual cortex~\cite{roy2018does, rokszin2010visual}. This design builds upon foundational studies of biological motion perception, including the Hassenstein--Reichardt detector~\cite{hassenstein1956systemtheoretische} and later extensions by Borst and Egelhaaf~\cite{borst1987temporal}, which established the principles of correlation-based Elementary Motion Detectors (EMDs).
In the proposed architecture, DS populations are implemented using spiking Elementary Motion Detectors (sEMDs) available within the SpiNNaker framework~\cite{milde2018spiking, clark2011defining, mauss2015neural}. Each sEMD incorporates a Time-Difference Encoder (TDE)~\cite{Chicca}, enabling the estimation of motion through the temporal relationship between neighbouring receptive field activations.
A TDE unit receives inputs from two adjacent RFs and encodes their relative activation timing, which is inversely related to the velocity of the moving stimulus. Following the architecture proposed in~\cite{Chicca}, one input is routed through a \textit{facilitation synapse} and the other through a trigger synapse. The facilitation synapse generates an exponentially decaying gain signal, while the trigger synapse produces an excitatory postsynaptic current whose amplitude is modulated by this gain. When a stimulus moves in the preferred direction, the facilitation input precedes the trigger input, increasing the postsynaptic response sufficiently to drive the membrane potential above threshold and elicit one or more spikes (Figure~\ref{fig:semd}). Within each $2\times 2$ neighbourhood, eight DS populations are defined to cover all possible motion pathways: two directions for each of the four orientations (horizontal, vertical, and the two diagonals). The facilitator–trigger ordering of the synaptic connections determines the preferred direction of each population (Figure~\ref{fig:directions}). The eight DS populations operate in parallel and feed a winner-take-all (WTA) population. Through lateral inhibition, the WTA suppresses less active populations, allowing the most active DS population to indicate the dominant direction of motion (see Readout Populations in Figure~\ref{fig:closed-loop}). As the TDE mechanism encodes traversal time through spike timing, the mean firing rate (MFR) of the winning population is expected to correlate with the object's velocity; this relationship is characterised in Section~\ref{ch:syst_charact}. 

\subsubsection{Angle readout population}\label{ch:AngleDetectionPopulation}
%% 3rd stage-4: SNN- Angle specificity
%36 neurons each representing 10 degrees to have a better angular specificity (rather than only 8 main directions with the resolution of 45 degrees)

The proposed model addresses the problem of estimating continuous motion direction directly on neuromorphic hardware from the eight DS populations introduced in the previous subsection. In several event-based vision systems, this decoding step is instead performed off-board using conventional processors~\cite{gallego2020event}. In this work, the goal is to perform the decoding fully on-board by introducing a dedicated Angle Readout (AR) population. To this end, the angle readout layer is composed of 36 neurons, each representing a preferred direction $\phi_j$ with an angular resolution of $10^\circ$. Each neuron in this population receives synaptic input from all neurons in the eight DS populations. A normalised Gaussian tuning function defines the connection strengths centred on the neuron's preferred angle. Specifically, the synaptic weight $w_{ij}$ from DS population $i$ to readout neuron $j$ in the AR population is defined by Equation~\ref{eq:w_angular}. This value is evenly divided among all individual synaptic connections from the $i$-th DS population to neuron $j$, such that their sum equals $w_{ij}$.

\begin{equation}
\label{eq:w_angular}
w_{ij} = \frac{\exp\left(-\frac{\mathrm{wrap}(\theta_i - \phi_j)^2}{2\sigma^2}\right)}
{\sum_k \exp\left(-\frac{\mathrm{wrap}(\theta_k - \phi_j)^2}{2\sigma^2}\right)}
\end{equation}

where $w_{ij}$ denotes the synaptic weight from the $i$-th DS population with preferred direction $\theta_i$ to the $j$-th angle readout neuron with preferred angle $\phi_j$. The operator $\mathrm{wrap}(\cdot)$ computes the angular distance on the unit circle, ensuring circular continuity. The parameter $\sigma$ controls the tuning width of the Gaussian, determining how sharply the contribution decreases as the angular difference increases. The normalisation in the denominator is performed over all DS populations indexed by $k$, with $k \in \{0,\dots,7\}$ and $\theta_k = k \times 45^\circ$, ensuring that the weights for each readout neuron sum to one.
~(see Figure~\ref{fig:closed-loop}, Readout Populations). The activity of each angle readout neuron is given by
\begin{equation}
R_j = \sum_{i=1}^{8} r_i \, w_{ij},
\end{equation}
where $r_i$ denotes the spike count or firing rate of DS population $i$ over a fixed temporal window. As a result of this weighted integration, intermediate motion directions that are not explicitly represented by any single DS population emerge naturally from the combined contributions of neighbouring DS responses. Finally, to improve resolution and suppress spikes from irrelevant angular neurons, inhibitory connections are introduced to implement a winner-take-all (WTA) mechanism.

\subsection{sP output and strike policy}\label{ch:strikePolicy}
%% 4th stage of the closed loop: output integration, velocity vector estimation, action prediction

The final stage of the pipeline integrates the extracted motion information to generate an actionable flipper-control decision. The outputs of the eight DS populations and the angle readout layer provide a spiking representation of the ball's motion direction ($\phi$) and an estimate of its speed ($v$).
In parallel, the output of the filter-map (Section~\ref{ch:Den_pop}) provides a coarse but reliable estimate of the ball’s instantaneous location within the field of view $(x, y)$. The spatial structure of this filtered representation enables identification of the currently active region of the scene, allowing the system to track the ball’s position relative to the left and right flippers. In this formulation, by using the estimated state vector $(x, y, v, \phi)$, the decision-making \textit{strike policy} module (running on the host CPU alongside the simulator) predicts the next position of the ball and selects the flipper to actuate (Figure~\ref{fig:closed-loop}).

At each timestep, the next position is predicted with a constant-velocity motion model from the estimated state vector. Given the estimated state $(x_t, y_t, v_t, \phi_t)$ at time $t$, the predicted position is computed as:
\begin{equation}
\begin{aligned}
x_{t+1} &= x_t + v_x \Delta t, \\
y_{t+1} &= y_t + v_y \Delta t,
\end{aligned}
\label{eq:update_x_y}
\end{equation}
where the distance travelled over $\Delta t$ is $d = v \Delta t$, giving velocity components $v_x = \frac{d}{\Delta t}\cos(\phi)$ and $v_y = -\frac{d}{\Delta t}\sin(\phi)$. The negative vertical component sign represents the image coordinate system, where the origin is located in the top-left corner. Over a single decision window ($\Delta t = \Delta T = 16.7$~ms), gravity, friction and collisions change the velocity of the ball negligibly relative to its magnitude, so the per-step displacement is dominated by the current velocity. A constant-velocity model is therefore an adequate approximation at this horizon, as the prediction is refreshed each window as new estimates arrive.

The \textit{strike policy} is a deliberately minimal and transparent design; it triggers the corresponding flipper whenever the predicted ball position enters that flipper's strike region, with no learning or tuning of timing beyond this geometric condition. This choice is intentional, as the aim of this work is to evaluate the neuromorphic perception pipeline rather than to engineer an optimal controller. Coupling the spiking state estimate to the simplest possible decision rule ensures that closed-loop performance reflects the quality of perception rather than policy sophistication. The geometry of the strike region depends on the flipper physics and is specified per regime in Section~\ref{ch:closed-loop}, where the policy is evaluated under two flipper regimes against human players.

\section{sP system characterisation}\label{ch:syst_charact}
%% we have three main configurable parameters: ∆T, σ and RF size
%% The goal is to have good metric scores: direction prediction, speed estimation and angle detection
%% By varying values for ∆T and RF and then for σ, and evaluating the metrics, we find the optimal configuration of ∆T, σ and RF size

System characterisation is performed by systematically varying three key parameters, receptive field (RF) size, spike accumulation window ($\Delta T$), and angular tuning width ($\sigma$), across a controlled range of ball velocities.
The accurate estimation of the velocity vector $(x, y, v, \phi)$ requires reliable measurements of both speed ($v$) and direction ($\phi$). Motion estimation in the DS populations is inferred from the mean firing rate (MFR) of the recorded neural activity of the TDE cells. Consequently, the effects of RF size and $\Delta T$ on direction selectivity and speed estimation are first evaluated to identify the most effective DS population configuration. Angular estimation is then assessed from the angle readout (AR) population, which combines the outputs of the eight DS populations through a weighted tuning mechanism. Because the AR population depends on accurate directional responses from the DS layer, the parameter $\sigma$ is optimised only after selecting the best RF size and $\Delta T$ configuration for the DS populations. Overall, the characterisation process first identifies the RF size and $\Delta T$ that maximise direction detection and speed estimation performance, and subsequently determines the optimal $\sigma$ value for angular estimation.

\subsection{Direction selectivity and speed estimation}\label{ch:Speedestim}
%% Experimental Setup and scenarios

To identify an optimal configuration of $\Delta T$ and RF size for the DS populations, the minimal Pinball Game Simulator introduced in Section~\ref{ch:sim_recordings} is used (see Figure~\ref{fig:simulators}). Eight experiments are defined, each corresponding to a discrete motion direction ($0^\circ, 45^\circ, \dots, 315^\circ$), with ball speeds uniformly changed in the range 0--1000\,px/s, increasing and decreasing in increments of 100 px/s. Each speed is maintained for a 1-second interval, and the entire sequence is repeated three times (See Figure~\ref{fig:simulators}b). This controlled setup enables a precise comparison between the simulated ground-truth trajectories and the motion estimated by the spiking-based Pinball network, also facilitating parameter tuning and validation of the event-based sensing and preprocessing stages.

For each experiment, the RF size in the segmentation map is varied across five configurations ($3\times3$, $5\times5$, $7\times7$, $9\times9$, and $11\times11$ pixels). Larger RFs increase robustness to fast motion but reduce spatial precision, while smaller RFs improve localisation accuracy at the expense of sensitivity to high-speed displacements (see Section~\ref{ch:SegmentationRFs}). For each RF size, $\Delta T$ is swept over 16.7\, ms, 40\, ms, and 100\, ms with a fixed event-count threshold of $N_{\mathrm{evt}} = 500$ events, and the resulting accuracy in both direction and speed estimation is evaluated.
$N_{\mathrm{evt}}$ is selected based on the geometric characteristics of the task. The ball has a radius of 15 pixels within a $256\times256$ playfield, corresponding to roughly 1\% of the total area. 

After the Gaussian downsampling of the ROI to $64\times64$, the ball occupies a small, bounded region of fewer than 200 active cells. Considering event noise, specular reflections, and the tendency of a single pixel to generate multiple events over time, a conservative upper bound of 500 events is adopted. This ensures that sufficient motion information is accumulated under most conditions while avoiding excessive integration delays.
Overall, the evaluation consists of $8 \times 5 \times 3$ configurations, corresponding to eight motion directions, five RF sizes, and three $\Delta T$ values. For each configuration, direction detection performance is assessed by comparing the outputs of the eight DS neurons in the readout population (each corresponding to one DS population) with the ground-truth motion vectors generated by the simulator. Specifically, spikes from the eight DS neurons are accumulated over each $\Delta T$ window, and the winning neuron is defined as the one with the highest spike count. Its associated class is taken as the predicted motion direction, and accuracy is computed using Equation~\ref{eq:acc-dir}.

\begin{equation}
\label{eq:acc-dir}
    \text{Accuracy} = \frac{N_{\text{correct}}}{N_{\text{total}}}
\end{equation}
where $N_{\text{correct}}$ denotes the number of trials in which the predicted 
direction matches the ground-truth direction, and $N_{\text{total}}$ is the total 
number of trials. Table~\ref{tab:ref-size-delta-t} reports the average direction-perception accuracy across the eight principal directions. The full directional breakdown is provided in Tables~\ref{sup:tab:ref-size-delta-t-Downward}--\ref{sup:tab:ref-size-delta-t-TR-BL} in the Supplementary Material.

\begin{table}[H]
    \centering
        \caption{Direction detection accuracy (average over all directions ) for different receptive field sizes in the segmentation map and $\Delta T$ values, evaluated over ball speeds ranging from 0 to 1000~px/s.}
        \label{tab:ref-size-delta-t}

        \begin{adjustbox}{max width=1\textwidth}
            \begin{tabular}{*6c}
                 \toprule
                     \textbf{$\Delta$T  } &  Accuracy [\%]  & Accuracy [\%]  & Accuracy [\%]  & Accuracy [\%]  & Accuracy [\%]\\
                      (ms) & (RF-size = 3x3)& (RF-size = 5x5)& (RF-size = 7x7)& (RF-size = 9x9)& (RF-size = 11x11)\\
                 \midrule
                        
                        16.7 & 68.50 & \textbf{82.73} & 73.64 & 51.89 & 31.77\\
                        40 & 74.78 & \textbf{85.82} & 77.47 & 58.13 & 36.74\\
                        100 & 80.46 & \textbf{88.58} & 81.06 & 65.87 & 44.47\\
                        
                 \bottomrule
            \end{tabular}
        \end{adjustbox}
\end{table}

\begin{figure}[h!] % Refrain from using [H] placement until the text is final
    \centering
        \includegraphics[width=1\textwidth]    {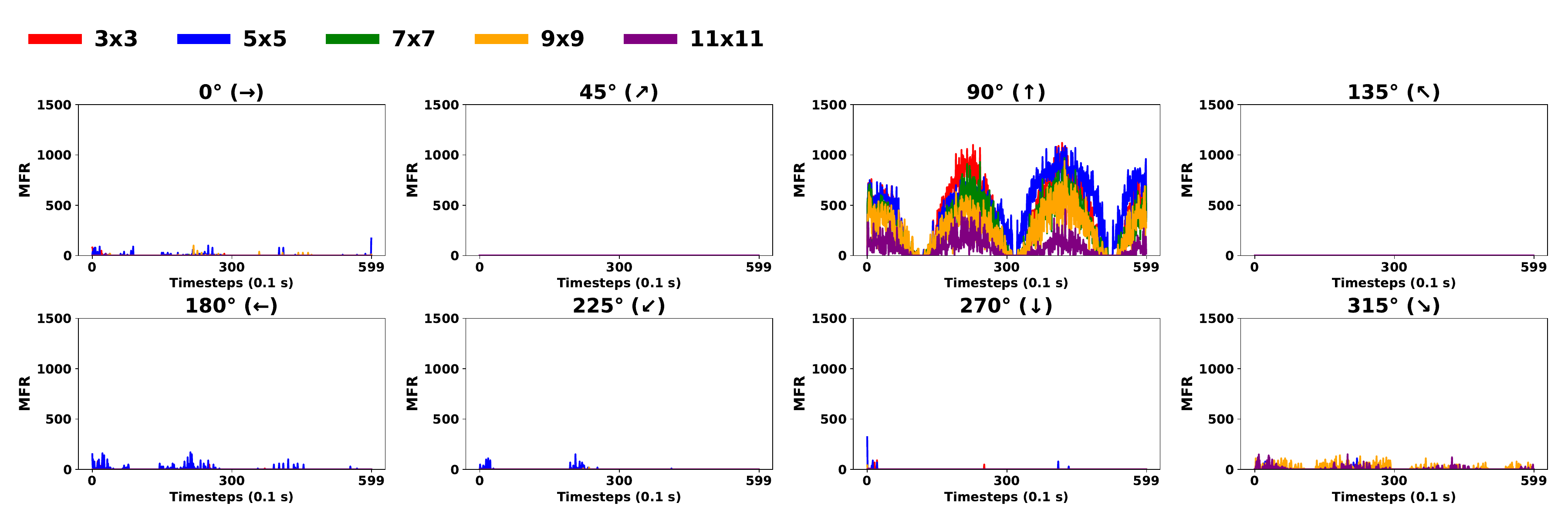}  
    \caption{Mean firing rate of the eight DS populations over a $100$~ms window. The population tuned to upward motion dominates, while other directions remain suppressed by the WTA mechanism.}
 
\label{fig:mfr_all}    
\end{figure}

%% Speed Estimation
Speed estimation in the proposed model is derived directly from the temporal dynamics of the TDE cells. As the ball traverses consecutive RFs, the TDE mechanism converts the time difference into characteristic spike bursts. Figure~\ref{fig:mfr_all} shows the MFR computed over a $100$~ms window for the eight DS populations. The neuron tuned to upward motion shows clear firing-rate modulation following changes in ball velocity, while the remaining populations remain largely inactive. This confirms the effectiveness of the WTA mechanism in suppressing non-relevant motion hypotheses and isolating a single dominant DS response.

%% Speed Estimation: Evaluation approach and results
As a result, the mean firing rate (MFR) of the winning DS neuron exhibits a monotonic relationship with object velocity, enabling speed estimation through temporal coding without explicit spatial tracking.
To quantify this relationship, both the MFR and the ground-truth speed are normalised prior to regression to remove scale differences and allow fair comparison across RF sizes and spike accumulation windows. Error metrics are therefore expressed in normalised units and capture estimation accuracy rather than absolute physical calibration (See Table~\ref{tab:mfr_speed_results}).

\begin{table*}[h]
\centering
\caption{Normalised speed estimation performance derived from the mean firing rate (MFR) of the winning DS population for different RF sizes and spike accumulation windows $\Delta T$. Both MFR and ground-truth speed are normalised prior to regression; error metrics are therefore reported in normalised units.}
\scriptsize
\setlength{\tabcolsep}{3pt}
\renewcommand{\arraystretch}{1.2}

\begin{tabular}{c|ccc|ccc|ccc|ccc|ccc}
\hline
% \multirow{2}{*}{\textbf{Metric}} 
& \multicolumn{15}{c}{\textbf{RF Size}} \\
\cline{2-16}

& \multicolumn{3}{c}{3$\times$3}
& \multicolumn{3}{c}{5$\times$5}
& \multicolumn{3}{c}{7$\times$7}
& \multicolumn{3}{c}{9$\times$9}
& \multicolumn{3}{c}{11$\times$11} \\
\cline{2-16}

\textbf{$\Delta T$ (ms)}
& 16.7 & 40 & 100
& 16.7 & 40 & 100
& 16.7 & 40 & 100
& 16.7 & 40 & 100
& 16.7 & 40 & 100 \\
\hline

\textbf{$r$
}& 0.873 & 0.906 & 0.936
& 0.871 & 0.902 & 0.931
& 0.788 & 0.827 & 0.895
& 0.715 & 0.779 & 0.852
& 0.540 & 0.603 & 0.725 \\

\textbf{$R^2$}
& 0.761 & 0.821 & 0.876
& 0.758 & 0.813 & 0.866
& 0.621 & 0.683 & 0.802
& 0.511 & 0.606 & 0.726
& 0.291 & 0.364 & 0.525 \\

\textbf{MAE}
& 0.374 & 0.338 & 0.291
& 0.396 & 0.348 & 0.283
& 0.471 & 0.422 & 0.349
& 0.575 & 0.516 & 0.419
& 0.702 & 0.659 & 0.571 \\

\textbf{RMSE}
& 0.488 & 0.423 & 0.353
& 0.491 & 0.432 & 0.366
& 0.616 & 0.563 & 0.445
& 0.699 & 0.628 & 0.523
& 0.842 & 0.798 & 0.689 \\

\hline
\end{tabular}

\label{tab:mfr_speed_results}
\end{table*}

\begin{figure}[h!] % Refrain from using [H] placement until the text is final
    \centering
        \includegraphics[width=0.95\textwidth]{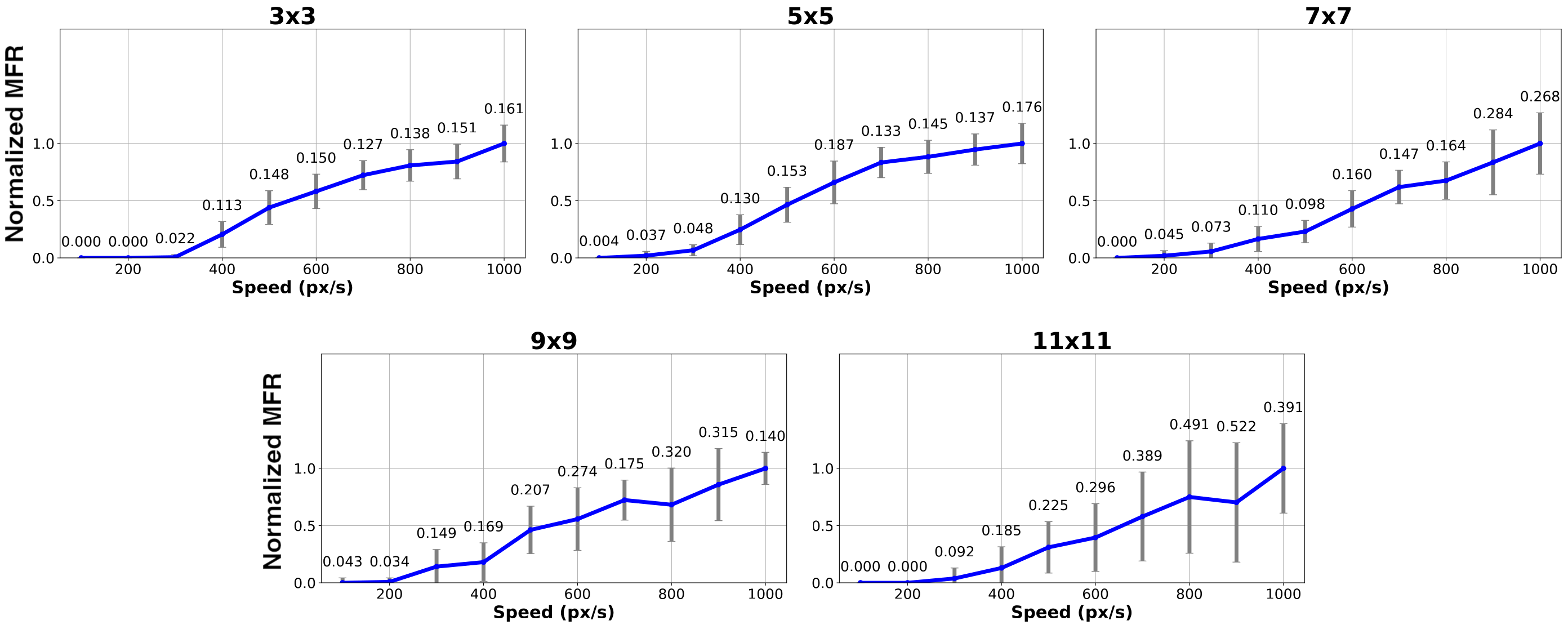}%{tracker_imgs/std_0_to_1000_upward_all_RF_1_row_0.1_ms.pdf}
    \caption{Standard deviation of the normalised mean firing rate (MFR) as a function of speed for different RF sizes.}
\label{fig:std_all}    
\end{figure}

%% Speed Estimation: Metrics explanation
Performance is evaluated using complementary metrics. The Pearson correlation coefficient ($r$) measures linear dependence between estimated and true speeds, while the coefficient of determination ($R^2$) quantifies the proportion of explained variance. Mean Absolute Error (MAE) reflects average estimation error, and RMSE emphasises larger deviations and outliers. 
Additionally, Figure~\ref{fig:std_all} reports the standard deviation of the normalised firing rate as a function of speed for different RF sizes.

%% Results analysis from Direction Detection Accuracy and Speed Estimation to find the optimal ∆T and RF size
Table~\ref{tab:ref-size-delta-t} and~\ref{tab:mfr_speed_results} show that both RF size and accumulation window $\Delta T$ significantly affect performance. Increasing $\Delta T$ improves all metrics by stabilising firing-rate estimates through longer temporal integration. The $\Delta T = 100$~ms condition yields the highest accuracy in Table~\ref{tab:ref-size-delta-t}, as well as the best correlation and lowest error across RF sizes, representing an upper bound on estimation performance.
However, this latency is not compatible with real-time pinball interaction. As discussed in Section~\ref{ch:dyn_event_acc}, larger $\Delta T$ also leads to greater inter-frame displacement of the ball, which degrades spatial consistency and positional accuracy. In contrast, shorter windows ($16.7$--$40$~ms) preserve temporal responsiveness while maintaining acceptable accuracy, particularly for smaller RFs ($3\times3$ and $5\times5$), which provide the best trade-off between precision and robustness. Moreover, smaller RFs exhibit lower and more stable variability across most speeds, whereas larger RFs show increased variance (Figure~\ref{fig:std_all}), especially at higher velocities. This behaviour is consistent with the degradation in estimation accuracy observed for larger RF configurations.
Overall, to ensure real-time operation and low-latency decision-making, the RF sizes of $3\times3$ and $5\times5$ combined with $\Delta T = 16.7$~ms are selected as the preferred operating configuration to investigate the angular specificity of the model in the following section.

\subsection{Angle specificity}\label{ch:AngleSpec}
%% angle specificity analysis using the optimal configuration from the previous section (∆T = 16.7 ms and RF_size = 3x3 and 5x5)

\begin{figure}[t]
  \centering
  \includegraphics[width=16cm, height = 8cm]{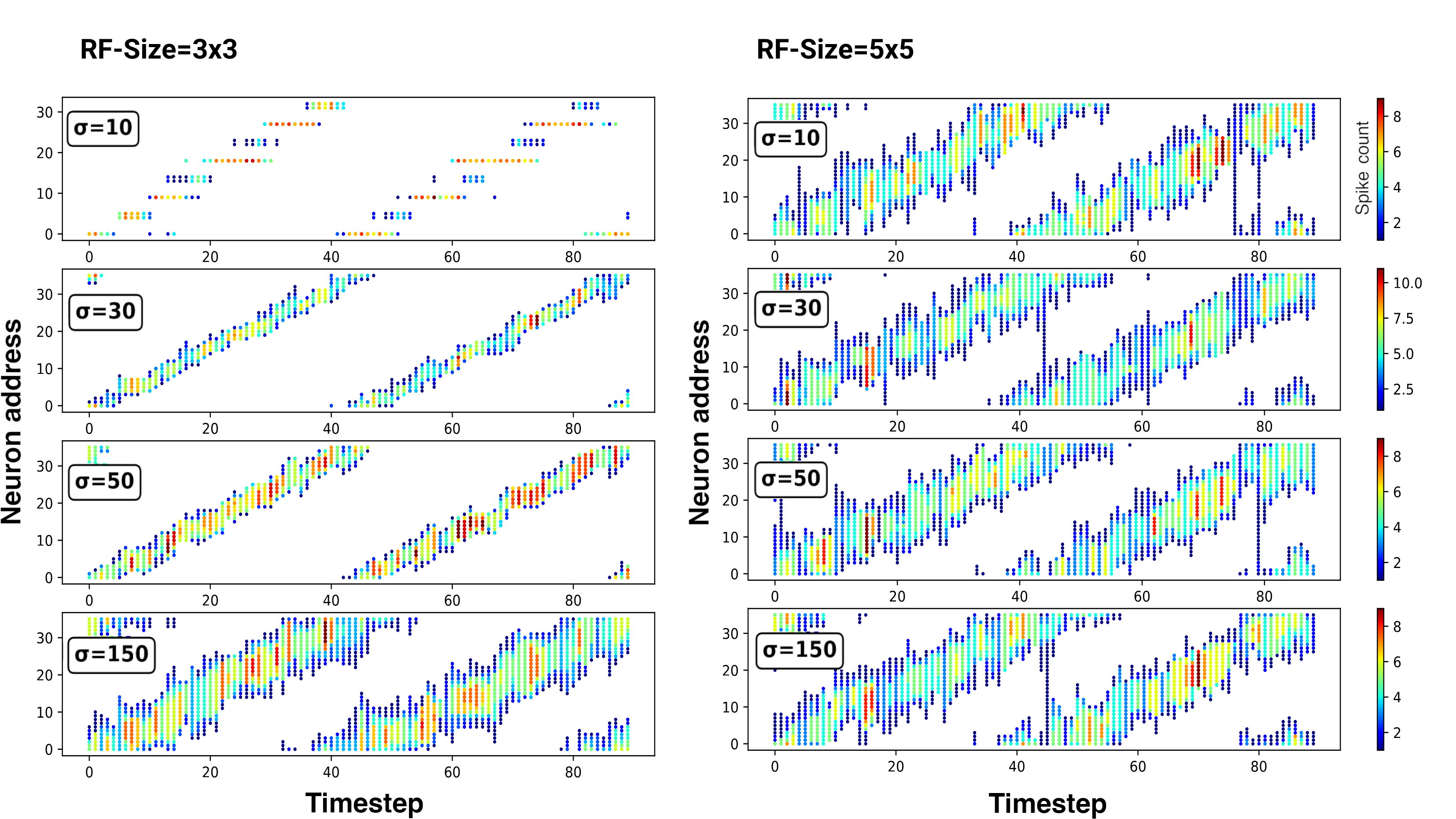}
\caption{Activity of the AR population during circular ball motion for different Gaussian tuning widths $\sigma$ and the RF size of $3\times3$ and $5\times5$.Neurons are ordered by preferred angle ($10^\circ$ resolution) and plotted over time. While a larger RF introduces excessive overlap, the RF size of $3\times3$ provides a good angular specificity. Smaller $\sigma$ values yield highly selective but sparse responses, while larger $\sigma$ values produce broader activation with reduced angular specificity. A moderate tuning width ($\sigma=30$) provides a balance between selectivity and smooth continuous angle representation.}
  \label{fig:circular}
\end{figure}

Beyond discrete direction classification, the angle readout (AR) population (See Figure~\ref{fig:closed-loop}) decodes a continuous estimate of motion orientation at $10^\circ$ resolution, providing finer angular precision than the eight DS populations alone (Further details in Section~\ref{ch:AngleDetectionPopulation}). Figure~\ref{fig:circular} shows the activity of the AR population for RF sizes of $3\times3$ and $5\times5$, evaluated on a circular ball trajectory spanning the full $0^\circ$--$360^\circ$ range at a constant speed of $1000$~px/s. The angular selectivity was evaluated at the maximum ball speed due to its importance for predicting the next ball position, as discussed in Section~\ref{ch:closed-loop}. Figure~\ref{fig:circular} also compares the effect of the tuning width parameter $\sigma$ in~Equation~\ref{eq:w_angular}, which controls the spread of the Gaussian weight distribution around each preferred angle. From top to bottom, $\sigma$ is set to 10, 30, 50, and 150. As shown in the figure, when the RF size is $5\times5$ pixels, the responses of the AR neurons are very similar for all values of $\sigma$. In all cases, a wide overlap between neurons is observed, activating many neurons in the population simultaneously. 
Consequently, the population response becomes broadly distributed and does not provide good angle specificity. 

In contrast, with an RF size of $3\times3$ pixels and $\sigma = 10$, neuronal activity is largely confined to a single angle neuron and occurs mainly around the eight principal DS directions, limiting continuous angular representation. Increasing $\sigma$ broadens the tuning response, enabling smoother transitions and activating multiple neighbouring angle neurons. However, excessively large values ($\sigma = 150$) introduce excessive overlap and reduce angular specificity. To achieve a balance between selectivity and smooth angular interpolation across DS populations, a moderate tuning width of $\sigma=30$ is selected for this work.

\subsection{Parameters selection}
%% Concluding the analysis: Winner configuration = ∆T = 167.7 ms and RF_size = 3x3
Based on the characterisation results, a final network configuration was selected to provide accurate speed estimation and direction perception while satisfying the real-time constraints of the pinball task.
The speed and direction detection capabilities of the model were evaluated across different RF sizes and accumulation windows. As shown in Section~\ref{ch:Speedestim}, increasing $\Delta T$ improves performance by stabilising neural firing rates. However, because real-time ball localisation and rapid flipper actuation are essential, the $\Delta T = 16.7$~ms is selected despite its slightly lower estimation accuracy. Using this accumulation window, the RF size analysis identified $3\times3$ and $5\times5$ as the most effective configurations for the DS populations. The subsequent angular specificity analysis showed that only the $3\times3$ RF provides sufficient angular resolution while maintaining limited overlap between neighbouring angle estimates. Furthermore, among the tested tuning widths, $\sigma = 30$ produced the best balance between smooth angular representation and angular specificity. Consequently, the final configuration adopted for the SpiNNaker implementation uses an RF size of $3\times3$, an accumulation window of $\Delta T = 16.7$~ms, and an angular tuning width of $\sigma = 30$. Parameters for the \acs{sP} network are shown in Table~\ref{tab:deployment_params}. Overall, these results demonstrate that compact RFs combined with short accumulation windows enable reliable motion estimation with low latency, making the proposed architecture well suited for real-time neuromorphic pinball tracking and control. 

% A good spot to add a table with complete parameter overview for the deployment
\begin{table}[H]
\centering
\caption{Deployment parameters for real-time operation of the system.}
\label{tab:deployment_params}
\footnotesize
\setlength{\tabcolsep}{5pt}
\renewcommand{\arraystretch}{1.1}
\begin{tabular}{@{}ll@{\hspace{1.5em}}ll@{\hspace{1.5em}}ll@{}}
\toprule
\multicolumn{2}{@{}l}{\textbf{Input \& denoising}}
& \multicolumn{2}{@{}l}{\textbf{Direction selectivity}}
& \multicolumn{2}{@{}l}{\textbf{Angle readout \& deployment}} \\
\midrule
Event camera      & EVK4-HD ($1280\times720$)          & RF size            & $3\times3$                  & Readout neurons & 36 ($10^\circ$) \\
ROI               & $128\times128$ px                  & DS populations     & 8 ($2\times2$, $45^\circ$)  & Tuning width    & $\sigma = 30$ \\
Camera distance   & $\sim$75 cm                        & Facilitation delay & $D_{\mathrm{fac}} = 2$ ms   & Platform        & SpiNNaker-5 \\
Monitor refresh rate      & 60 Hz                              & Trigger delay      & $D_{\mathrm{trig}} = 1$ ms  & Total neurons   & 24{,}620 \\
AECP accumulation      & $N_{\mathrm{evt}}=500$, $16.7$ ms  &  & & &  \\
Event polarity    & positive only                      &    & &                 & \\
Gaussian kernel    & $3\times3$, stride $(2,2)$              &      &  &                 & \\
 Kernel width       & $\sigma_{\mathrm{G}} = 0.8$     &       & &                 & \\
\bottomrule
\end{tabular}
\end{table}

\section{Experiments and Results: closed-loop performance evaluation}\label{ch:closed-loop}
%% The full pipeline in action: The results of game paying. Defining a metric of hitting the ball showing the model capability in 1- detectting the current position of the ball, 2-calculating the velocity vector and 3-predicting the next position of the ball. Also it include the full pipeline latency and model size of the SNN.

After characterising the model and selecting the optimal parameter configuration, the full system is integrated to perform closed-loop pinball gameplay (See Figure~\ref{fig:closed-loop}). The DVS camera captures the pinball scene of the \textit{interactive game simulator} (Section~\ref{ch:sim_recordings}, Figure~\ref{fig:simulators}), and a $128\times128$ pixel ROI centred on the flipper and drain area is transmitted to the SpiNNaker system for real-time processing.
Within the system, spatial information is extracted from the filter-map population (See Figure~\ref{fig:closed-loop}), providing an estimate of the ball’s instantaneous position $(x, y)$. Motion direction and orientation are decoded from the DS and AR populations, while ball speed $v$ is estimated from the mean firing rate (MFR) of the winning DS neuron using the regression model described in
Section~\ref{ch:Speedestim}. All features are computed over a decision window of $\Delta T = 16.7$~ms, corresponding to a $60$~Hz control rate. At each time step, the next ball position is predicted using a constant-velocity motion model from the estimated state vector $(x, y, v, \phi)$, as described in Section~\ref{ch:strikePolicy}. 

Based on these predictions, the strike policy (Section~\ref{ch:strikePolicy}) sends a control command to the game simulator to activate the appropriate flipper, triggering it whenever the ball is predicted to enter that flipper's strike region. The geometry of this region is specified separately for each experiment. Together, the neuromorphic perception pipeline, the strike policy, and the flipper control signal constitute a single closed-loop agent, which we refer to as the neuromorphic-perception-to-action (\acs{nPA}) loop throughout this section. The closed loop is evaluated in simulation in two regimes: a binary-gate flipper regime (Section~\ref{ch:BinaryFlip}) and a solid flipper regime with more realistic dynamics (Section~\ref{ch:SolidFlip}).
Finally, the operation of the loop is demonstrated on a physical pinball system (see Section~\ref{sec:physicalPinball}). 

\subsection{Binary gate flipper regime}\label{ch:BinaryFlip}
In the binary flipper regime, the flippers act as momentary action-gated colliders. A flipper exists as a collidable surface only during the frame in which it is actively triggered: a correctly timed trigger deflects the ball back into play, while a mistimed one leaves no collider present, and the ball passes through the flipper area along its original trajectory. Since the flipper has no physical presence unless correctly triggered, the interaction reduces to a single all-or-nothing timing decision.
The strike region is a disc of one ball-radius around the flipper: the reactive policy triggers when the predicted ball position falls within this region, while a hit is scored when the actual ball position satisfies the same condition. Pairing the binary gate with a triggering rule matched to the hit-success criterion isolates the perception-and-prediction chain as far as possible, excluding post-trigger physics, so a miss reflects perception and prediction accuracy rather than the trigger threshold.

To evaluate the \acs{nPA} loop performance, number of successful hits and flipper activations is counted. A \textit{hit} is recorded when a flipper is activated and successfully strikes the ball. A \textit{miss} is recorded when a flipper is activated but fails to hit the ball, either due to inaccurate timing, incorrect flipper selection, or false detection. Let $N_{\text{hit}}$, $N_{\text{miss}}$ and $N_{\text{act}} = N_{\text{hit}} + N_{\text{miss}}$ denote their totals over $N_{\text{ep}}=100$ episodes (one ball each). From these, two complementary metrics are reported: \textit{hit rate} and \textit{miss rate}.

\begin{equation}
\label{eq:hit_rate}
\mathrm{Hit\ Rate} = \frac{N_{\text{hit}}}{N_{\text{act}}},
\end{equation}

\begin{equation}
\label{eq:miss_rate}
\mathrm{Miss\ Rate} = \frac{N_{\text{miss}}}{N_{\text{act}}}.
\end{equation}

While these rates quantify the accuracy of each activation, as ratios, they are independent of the number of activations performed: a precise and an aggressive player may achieve identical hit rates despite markedly different behaviour. To capture this aspect of behaviour, the economy of control is additionally reported as two per-episode counts,
\begin{equation}
\label{eq:per_episode}
\frac{N_{\text{hit}}}{N_{\text{ep}}}, \qquad \frac{N_{\text{act}}}{N_{\text{ep}}},
\end{equation}
the hits achieved and the actions spent to achieve them. The hit rate is their ratio; reporting the counts alongside it distinguishes a system that scores from one that merely acts often. These metrics directly reflect the effectiveness of the \acs{nPA} loop and quantify the system’s ability to translate spiking sensory information into accurate and timely motor decisions.

To provide a baseline for evaluating the efficiency of the loop and to assess how its real-time performance compares to human reaction capabilities, the \acs{nPA} is compared against human players on the identical task. This evaluation conducts a user study in which 10 participants played the pinball game using manual keyboard control of the flippers\footnote{\url{https://github.com/MazdakFatahi/PinBallSimulator}}. Each participant played 100 balls, with the maximum ball speed limited to 1000~px/s. The game is a physics-based simulator in which parameters such as gravity, continuous friction, wall and flipper reflection coefficients, and energy loss during wall bounces dynamically affect the ball’s speed throughout the simulation. As a result accurate action time is not easy to predict, and both the number of flipper activations and the number of successful hits vary across players (Table~\ref{tab:human_test}). The \acs{nPA} loop was evaluated over 10 independent runs using the same visual setup and ball count, the average over these runs is reported in the last column of Table~\ref{tab:human_test}.

\begin{table}[h]
\centering
\caption{Binary gate flipper regime: Human baseline performance in the pinball game compared with the average results of the  \acs{nPA} loop.}
\setlength{\tabcolsep}{4pt}   % column spacing (default ~6pt)
\renewcommand{\arraystretch}{1.0} % row spacing (default 1.2)
\begin{tabular}{lcccccccccccccc}
\hline
 & \multicolumn{10}{c}{Players} &  &  &  & \acs{nPA}$^{*}$  \\
\cline{2-11}
Metric 
& 1 & 2 & 3 & 4 & 5 & 6 & 7 & 8 & 9 & 10 & Min & Max & Avg & Avg \\
\hline
Successful hits 
& 28 & 34 & 50 & 49 & 33 & 40 & 127 & 112 & 62 & 48 
& 28 & 127 & 58.3 &\textbf{ 64} \\

Number of actions 
& 321 & 56 & 317 & 422 & 197 & 88 & 436 & 394 & 384 & 206 
& 56 & 436 & 282.1 &\textbf{114} \\

Hit rate (\%) 
& 9.0 & 61.0 & 16.0 & 12.0 & 17.0 & 45.0 & 29.0 & 28.0 & 16.0 & 23.0 
& 9.0 & \textbf{61.0} & \textbf{25.6} & \textbf{56.1} \\

Miss rate (\%) 
& 91.0 & 39.0 & 84.0 & 88.0 & 83.0 & 55.0 & 71.0 & 72.0 & 84.0 & 77.0 
& 39.0 & 91.0 & 74.4 & \textbf{43.9} \\
\hline
% \begin{flushleft}
$^{*}$ \footnotesize Neuromorphic-Perception-to-Action
% \end{flushleft}
\end{tabular}
\label{tab:human_test}
\end{table}

\begin{figure}[htbp]
  \centering
  % \fbox{
    \includegraphics[width=.6\textwidth]{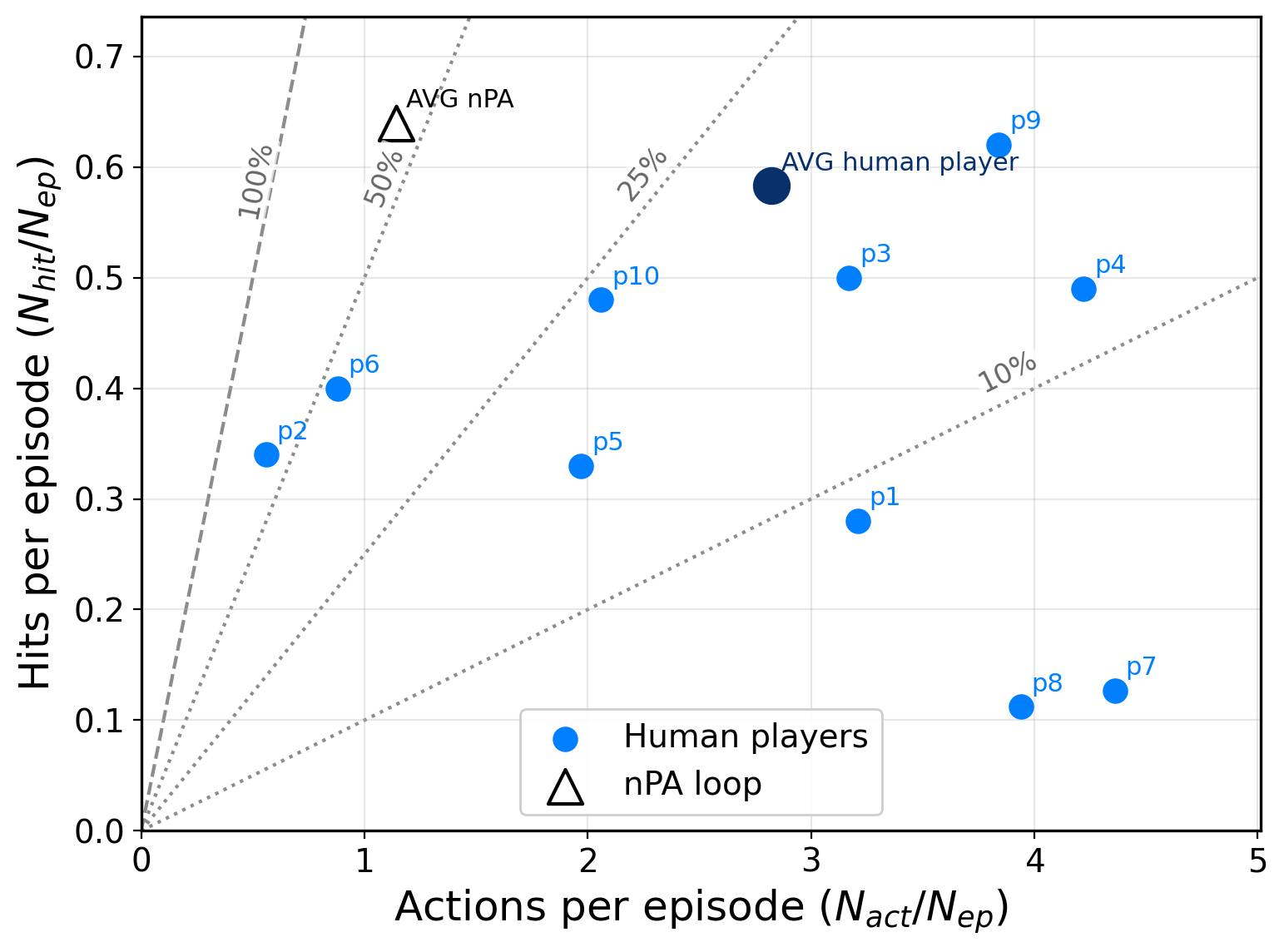}
  % }
\caption{Binary gate flipper regime: Human baseline performance in the pinball game, compared with the average result of the proposed pipeline.
}
\label{fig:exp1Economy}
\end{figure}

Figure~\ref{fig:exp1Economy} reports hits and actions per episode for each participant and for the \acs{nPA} loop. Results of human players are scattered across both axes, tracing the task's accuracy–economy trade-off. Participants adopted strategies from precise (player~2 and~6 achieving high hit rates from few actions) to aggressive (players~4 and~7, acting several times as often for comparable return). The \acs{nPA} loop occupies the favourable upper-left corner, matching the top of the human hit range while spending far fewer actions than any human of comparable output. Equivalently, its 56.1\% hit rate is comparable to the most accurate individuals and more than double the human average of 25.6\% (Table~\ref{tab:human_test}). 
This economy is diagnostic of perception quality: a high hit count at low action counts implies estimated position and velocity accurate enough for reliable strike timing under real-time constraints, whereas noisier perception would require many activations for the same return, shifting an agent rightward, as seen for the more aggressive players. Achieving this at a fixed rate of one command per $\Delta T = 16.7$~ms indicates that fully spike-based event-driven perception can operate within the latency budget required for closed-loop sensorimotor control; the corresponding latency and energy numbers are reported in Section~\ref{ch:Efficiency}.

\subsection{Solid flipper regime}\label{ch:SolidFlip}
The binary-gate regime of Section~\ref{ch:BinaryFlip} isolates the perception–action chain, but at the cost of physical realism, penalising human players, whose intuition of the game is rooted in its physical dynamics. To provide a fairer comparison, the system has been evaluated by introducing a solid flipper regime in which the flippers are persistent colliders, occupying the playfield at all times and deflecting the ball on contact, whether or not they are actively triggered. 
In this scenario, the strike region is the triangular area swept by the flipper. As before, the policy triggers when the predicted ball position overlaps it, and a hit is scored when the actual ball position satisfies the same condition. Alongside hit rate and per-episode action economy, the \textit{conversion rate} is reported, defined as the fraction of confirmed ball–flipper contacts converted into active strikes rather than passive bounces.

\begin{equation}\label{Eq:conversion}
 \text{Conversion rate} = \frac{N_{\text{hit}}}{N_{\text{hit}}+ N_{\text{bounce}}}
\end{equation}

As in the binary regime, the \acs{nPA} loop is compared with human players on the identical task. Twenty participants played the solid flipper game using manual keyboard control, each completing $N_\mathrm{ep}=30$ balls under the same visual setup and ball-speed limit as in the Binary gate flipper regime (See details in Section~\ref{ch:BinaryFlip}). To prevent players from simply holding the flippers permanently active, each flipper was limited to a maximum activation rate of once per 10 frames. To see how player behaviour adapts to intent, participants were split into two groups of ten, each given a different instruction. The first group was asked to trigger a flipper only when confident of a hit, prioritising precision over volume (\emph{avoid misclicks}); the other was asked to press liberally, prioritising the number of successful hits over the risk of wasted activations (\emph{maximise score}). This yields two contrasting human strategies against which to position the system, rather than a single undifferentiated baseline.

 \begin{figure}[htbp]
  \centering
  % \fbox{
    \includegraphics[width=\textwidth]{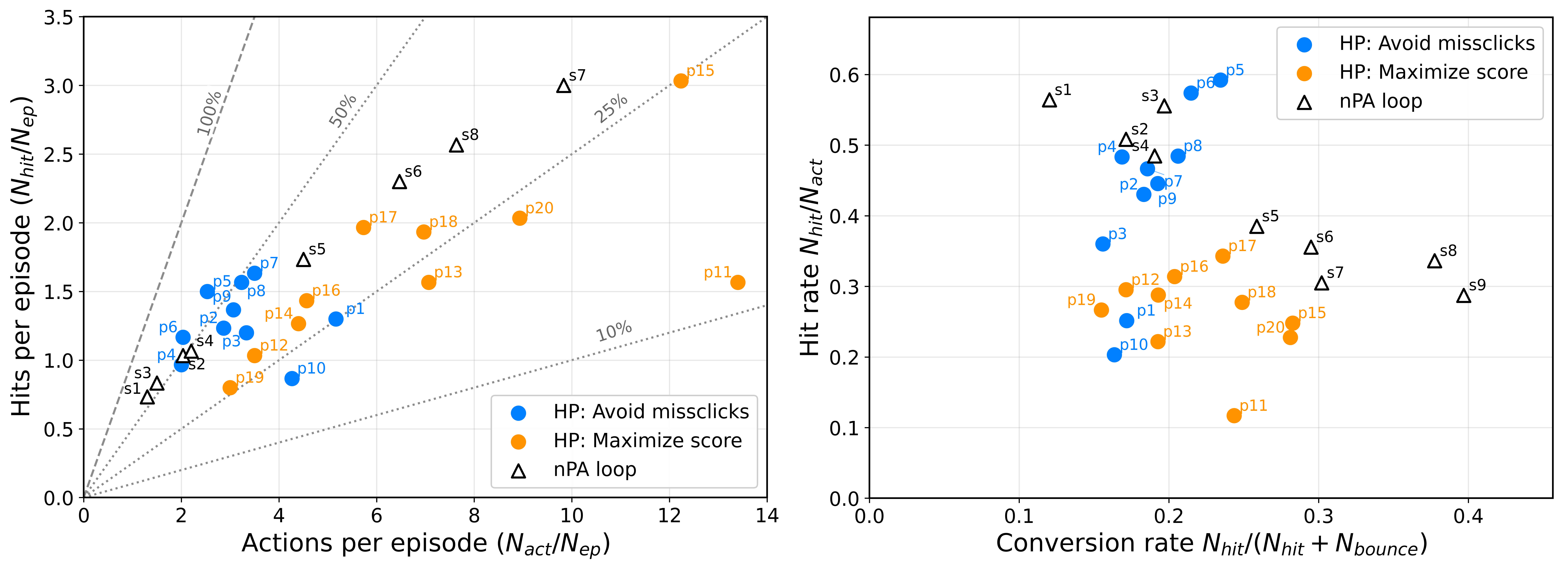}
  % }
\caption{Solid flipper regime: human player (HP) baseline against the \acs{nPA} loop swept across strike-region sizes ($s_1$–$s_9$, smallest to largest). Action economy (left) and precision–conversion trade-off (right). A small strike region triggers only on balls predicted to pass close to the flipper, yielding few but well justified activations, while a larger region admits more marginal predictions, raising activation count at the cost of precision.}
\label{fig:exp2Economy}
\end{figure}

Figure~\ref{fig:exp2Economy} reports solid flipper regime performance as per-episode action economy (left) and the complementary precision-conversion trade-off (right).Rather than characterising the \acs{nPA} loop by a single operating point, the size of its strike region is swept, varying how readily the policy commits to a strike. Sweeping from the smallest to the largest region ($s_1-s_9$) traces a controlled progression from precise to aggressive play, exposing the loop as a tunable controller capable of reproducing the full range of observed human strategies.
This range is set by the two instruction groups, which adopt sharply different strategies. In the economy plot (left), the avoid-misclicks group stays in the low-action region, while the maximise-score group spends far more freely; the precision–conversion view (right) separates them along hit rate, from $\sim$0.6 for the most precise avoid-misclicks players down to $\sim$0.2–0.35 for the maximise-score group.
By sweeping the strike region alone, the \acs{nPA} loop can be driven to perform across the human range; small strike regions ($s_1–s_4$) place it alongside the most precise players, while enlarging the region moves it through the mid-range ($s_5–s_6$) into the aggressive, high-action strategy ($s_7–s_9$).

Notably, this coverage holds in the solid flipper regime, where realistic contact physics remove the disadvantage the binary gate imposed on human players. Across the sweep, the \acs{nPA} loop spans the human player range in both action economy and hit rate, and reaches comparable conversion rates over confirmed ball–flipper contacts. 
Since conversion is measured only on balls that reach a flipper, each converted contact requires accurate localisation and arrival prediction to strike in time. The competitive conversion rates across the sweep therefore indicate that the spiking perception–prediction chain remains reliable even under the fairer, physically realistic condition.

We also demonstrated the pipeline with a DAVIS346 MONO camera to show that it is not tied to a specific event-camera model. However, this was only a compatibility demonstration. All experiments and quantitative results presented in this work were obtained with the Prophesee camera.

\section{Pinball demonstrator}\label{sec:physicalPinball}
\begin{figure}[!htbp]
  \centering
  % \fbox{
    \includegraphics[width=\textwidth]{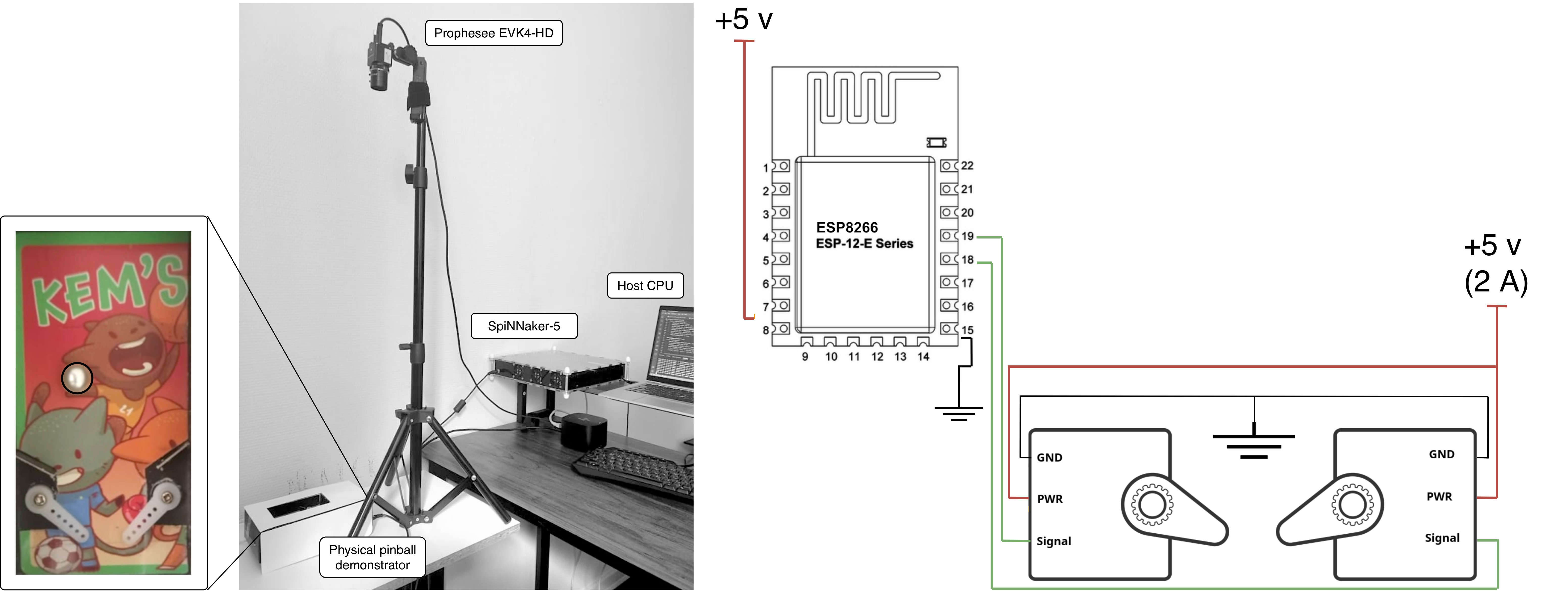}
  % }
\caption{Physical pinball demonstrator. \textbf{Left:} Schema of the setup: a metal ball in a box with two servo-actuated flippers. \textbf{Right:} the actuation wiring, in which an ESP8266 module drives the two flipper servos, receiving trigger commands over
Wi-Fi from the same control pipeline used for the simulated game. The playfield is observed by the Prophesee DVS and controlled through the neuromorphic-perception-to-action (\acs{nPA}) loop.}
\label{fig:physicalpinball}
\end{figure}

The preceding experiments were conducted on a simulated game environment to allow controlled, repeatable performance measurements. To assess whether the same principle operates outside simulation, the physical pinball demonstrator has been additionally built (See Figure~\ref{fig:physicalpinball}), in which a metal ball moves within a box, and two flippers are driven by servo motors. Flipper actuation is handled by an ESP8266 module driving two servo motors, one per flipper (Figure~\ref{fig:physicalpinball}, right). Each servo switches between two preset angles corresponding to the flipper's raised and lowered states. The ESP8266 receives its trigger
commands from the same processing module that controls the simulated game; rather than actuating virtual flippers, the commands are transmitted over a Wi-Fi socket to the ESP8266. The perception and decision stages are thus reused unchanged, and only the final actuation
step is redirected from the simulator to the physical hardware. Under this configuration, the \acs{nPA} loop successfully tracks the metal ball and actuates the physical flippers in closed loop, keeping the ball in play. A supporting video of the demonstrator in operation is provided as a proof of concept\footnote{Video available at:  \href{https://youtu.be/2iy6tDeiyB8}{https://youtu.be/2iy6tDeiyB8}}.

\subsection{System efficiency}\label{ch:Efficiency}

The efficiency of the deployed system is finally characterised along two axes relevant to real-time edge operation: end-to-end latency and dynamic power consumption.

\textbf{Latency:} The spiking \acs{sP} model runs asynchronously on the SpiNNaker-5 platform (SpiNN-5). The complete network consists of $\sim$25\,k neurons (as detailed in Table~\ref{tab:neuron_count}), while exhibiting an end-to-end input-output latency of approximately $5\,\mathrm{ms}$, following from the per-layer synaptic delays listed in Table~\ref{tab:synaptic_delays}.

%% decision making time= end-toend delay+accumulation time
Feature extraction for direction and speed additionally relies on an accumulation window of $16.7\,\mathrm{ms}$ (the accumulation upper bound) to form a firing-rate estimate from the population spike activity.
Due to the asynchronous and pipelined execution, early network layers can receive new sensory spikes, while later layers are still emitting output spikes for decision making. The effective decision latency is given by the sum of the accumulation window and the internal network latency, resulting in an overall reaction time of approximately $21.7\,\mathrm{ms}$. Once the pipeline is filled, a decision is produced for every event package; with the accumulation window bounded above by $\Delta T =16.7$ ms, the control rate is at least $60\,\mathrm{Hz}$ and increases under fast motion as the adaptive AECP event accumulation window shortens.

\begin{table}[!h]
\centering
\caption{Number of neurons in each population of the spiking-based Pinball (\acs{sP}) network.}
\label{tab:neuron_count}
\setlength{\tabcolsep}{4pt}   % column spacing (default ~6pt)
\renewcommand{\arraystretch}{1.0} % row spacing (default 1.2)
\small
\begin{tabular}{lcccccc}
\hline
Population 
& Input 
& Filter-Map 
& Segment-Map 
& DS
&  DS WTA
& Angle WTA \\
\hline
Number of neurons 
& 16,384 $(128\times128)$ 
& 4,096 $(64\times64)$ 
& 484 $(22\times22)$ 
& 3,612  
& 8 
& 36 \\
\hline
Total neurons 
& \multicolumn{6}{c}{24,620} \\
\hline
\end{tabular}
\end{table}

\begin{table}[!h]
\centering
\caption{Synaptic delays between successive layers in the neuromorphic \acs{sP} pipeline.}
\label{tab:synaptic_delays}
\setlength{\tabcolsep}{4pt}   % column spacing (default ~6pt)
\renewcommand{\arraystretch}{1.0} % row spacing (default 1.2)
\small
\begin{tabular}{lcccc}
\hline
Synaptic connection 
& Input--Filter-Map 
& Filter-Map--Segment-Map 
& Segment-Map--DS 
& DS--WTA / DS--Angle Readout \\
\hline
Delay (ms) 
& 1 
& 1 
& 2 {\small($D_{\mathrm{trig}}=1$, $D_{\mathrm{fac}}=2$)} 
& 1 \\
\hline
\end{tabular}
\end{table}

\textbf{Dynamic power:} Following established power models for SpiNNaker \cite{stromatias2013power, stromatias2015scalable}, the total power consumption of the board can be expressed as:
\begin{equation}
P_T = P_I + P_B + (P_N \times n) + (P_S \times s),
\end{equation}
where $P_I$ denotes the idle power of the board after booting, $P_B$ is the baseline power consumed by system services (SARK and the SpiNNaker API), $P_N$ is the power required to simulate a single LIF neuron with a 1~ms time-step, $n$ is the neuron count, $P_S$ is the energy consumed per synaptic event, and $s$ is the total number of synaptic events.  
Although this equation captures the overall power budget of the SpiNNaker system, the analysis focuses on the activity-dependent \emph{dynamic power}, $P_S \times s$, which isolates the energy spent generating spikes in response to events from
the fixed idle and baseline terms that do not depend on the network activity.

\begin{table}[h]
\centering
\caption{Approximate spike activity and corresponding dynamic energy consumption per population, assuming an average energy cost of 8~nJ per synaptic event. Reported values are rounded for clarity.}
\small
\begin{tabular}{lcc}
\toprule
\textbf{Population} & \textbf{Number of Spikes} & \textbf{Energy (nJ)} \\
\midrule
Filter-Map          & 101 & 808  \\
Segment-Map         & 8   & 64   \\
DS-Downward              & 2   & 16   \\
DS-Upward                & 32  & 256  \\
DS-Rightward             & 15  & 120  \\
DS-Leftward              & 11  & 88   \\
DS-Diagonal TL--BR       & 1   & 8    \\
DS-Diagonal BR--TL       & 16  & 128  \\
DS-Diagonal BL--TR       & 16  & 128  \\
DS-Diagonal TR--BL       & 1   & 8    \\
DS-WTA                & 15  & 120  \\
Angle Readout       & 91  & 728  \\
\midrule
\textbf{Total}        & \textbf{309} & \textbf{2,472} \\
\bottomrule
\end{tabular}
\label{tab:energy_population_breakdown}
\end{table}

Table~\ref{tab:energy_population_breakdown} reports the average spike counts per population over a $\Delta T = 16.7$~ms window, and the corresponding dynamic energy consumption, taking the established energy cost of 8~nJ per synaptic event \cite{stromatias2013power, stromatias2015scalable, romero2023high, van2018performance}.
The resulting average of 309 synaptic events per time window yields an estimated dynamic energy consumption of 2,472~nJ per $\Delta T$, corresponding to an average power consumption of about 148~$\mu$W. These values represent an \emph{upper bound}, as they were measured on the minimal game simulator (Section~\ref{ch:sim_recordings}), where the ball is continuously present in the field of view and drives spikes at every stage. In realistic gameplay, the ball leaves the ROI for extended periods, during which the network is largely silent and dynamic power is negligible; thus, actual consumption is lower than the reported one.

\section{Discussion}\label{ch:discussion}
%% Discussing the model efficiency including dynamic energy consumption analysis based on the.
%% NOTE: neuromorphic closed-loop (ncl) and spiking pipeline (sP) are not the same: ncl includes the game simulators and iteraction btw the board and CPU, but sP is only the SNN running on SpiNNaker

This work develops a fully neuromorphic motion-estimation (\acs{sP}) network deployed on the SpiNNaker platform as the perceptual core of a closed-loop pinball agent. The network drives a real-time perception-to-action loop directly from raw DVS events, sustaining the dynamic visuomotor task of pinball gameplay. Across two flipper regimes of the game (Section~\ref{ch:closed-loop}), the proposed neuromorphic-perception-to-action (\acs{nPA}) loop performed within the range spanned by human players while operating under a bounded real-time budget and at sub-milliwatt dynamic power.

%characterisation trade off
The deployed configuration itself reflects a deliberate trade-off investigated during characterisation. The highest direction-detection accuracy and the strongest speed-estimation correlations were both obtained with longer accumulation windows and, for direction, with $5\times5$ receptive fields (Section~\ref{ch:syst_charact}); however, these settings degrade angular specificity and are incompatible with the real-time constraints of closed-loop control. The configuration ultimately adopted, $3\times3$ receptive fields with $\Delta T = 16.7\,\mathrm{ms}$, sacrifices some raw accuracy for both angular resolution and latency. Despite operating away from the accuracy-maximising end of the characterisation sweep, the resulting perception remained reliable enough to sustain human-comparable, and at times human-exceeding, performance under real-time constraints, evidence that the reduced accuracy at this operating point did not translate into a meaningful loss of closed-loop capability.

% Sparse activation motion estimation
The pinball setting places a stringent demand on the perception stage. A small, fast ball activates only a sparse subset of receptive fields as it crosses the field of view, in contrast to the stimuli used in most prior EMD/TDE evaluations (wide-field bars, drifting gratings, or large moving objects that activate broad regions of the sensor and therefore probe only coarse directional selectivity ~\cite{giulioni2016event, d2020event, fschoepe2024finding}).
Under this sparser activation, the network nevertheless recovers a continuous $10^\circ$ angular estimate and a velocity vector, rather than a coarse eight-way direction. Sparse activation is the more demanding condition for a local motion detector, and it is also the one relevant to tracking a small target for downstream trajectory estimation, where the informative signal is confined to a few active cells.

% Human-comparable control under naive decision policy
A second consideration is how well this perception supports action. The strike policy is deliberately minimal; it commits a flipper whenever the predicted position enters the strike region, with no learned or tuned timing, so the closed-loop outcome depends primarily on the quality of the spiking state estimate rather than on the controller. This is reflected in the action economy of the loop: in the binary-gate regime, it achieved a hit count within the upper part of the human range while issuing far fewer activations than human players of comparable output (Table~\ref{tab:human_test}). A high hit count obtained from few activations is consistent with position and velocity estimates accurate enough to time individual strikes, whereas noisier estimates would require many speculative activations for the same return, as seen for the more aggressive human players. The loop's position on the accuracy--economy plane is therefore an indirect indicator of perception quality, although action economy alone does not isolate perception from task difficulty or the specific policy; comparison is made against the range of human performance rather than the group mean, since the small participant pool and varied engagement make the mean an unstable reference.

%solid-flipper regime
The solid-flipper regime extends this comparison under more realistic contact dynamics, where flippers persist as colliders rather than acting only at the instant of a trigger. Sweeping a single geometric parameter, the size of the strike region, moved the loop continuously across the full range of human strategies observed in the user study, from the low-action, high-precision behaviour of the avoid-misclicks group to the high-action, lower-precision behaviour of the maximise-score group, without any change to the perception pipeline. This is a stronger result than matching a single human operating point, it indicates that the spiking state estimate is accurate enough to support a whole family of viable strike policies, and that the trade-off between precision and activation volume, which for human players reflects a deliberate behavioural strategy, can be reproduced in the closed loop by tuning one interpretable parameter of an otherwise fixed controller. The competitive conversion rates maintained across this sweep further suggest that perception quality, not the decision rule, is what principally bounds performance in both regimes.

% Efficiency 
The behavioural results are obtained under real-time, low-power operation. The system runs within a bounded latency budget, with an internal network latency of about $5\,\mathrm{ms}$ and an effective reaction time near $21.7\,\mathrm{ms}$ once the accumulation window is included, at an estimated dynamic power of $\approx 148\,\mu\mathrm{W}$ using fewer than $25\,$k neurons (Section~\ref{ch:Efficiency}).
The reported power figure corresponds to a \emph{worst-case scenario}, in which the ball is always present within the field of view, continuously generating DVS events and forcing the neuromorphic pipeline to emit spikes at all stages. This scenario was deliberately chosen to enable a consistent and interpretable estimation of energy consumption. In realistic gameplay conditions, the ball is absent from the region of interest for significant periods, during which the network remains largely inactive and consumes negligible dynamic energy. Therefore, the reported energy consumption should be interpreted as an upper bound.
SpiNNaker is not the lowest-power neuromorphic platform available. Still, its programmability and low-latency spike routing suit a pipeline of this complexity, and its many-core fabric avoids the memory bottleneck of GPU-based implementations~\cite{davison2009pynn, furber2014spinnaker}. 
Table~\ref{tab:toy_examples} situates these figures within the broader landscape of event-based, neuromorphic closed-loop demonstrators. Among the small number of systems that combine spiking perception with neuromorphic computing (See Table~\ref{tab:toy_examples}, bold examples), the proposed pipeline reports the lowest dynamic power consumption. Direct latency comparison is inherently difficult, as reaction time depends strongly on the dynamics of the game and the specific network design rather than on hardware alone. The only reported latency lower than that of the proposed system is achieved by the table tennis demonstrator of Ziegler et al.~\cite{ziegler_tennis} on Akida; however, that system reports only ball position and does not perform velocity estimation. A further interesting latency figure is reported by Gava et al.~\cite{Gava_PUCK}, at $8\,\mu\mathrm{s}$ on a CPU Intel i7; however, this value measures how far behind real time the algorithm's internal event representation is at the moment of each update, rather than the time required to produce a full position or velocity estimate, and is therefore not directly comparable to the closed-loop reaction time reported here.
Among the systems that combine an event-based camera for perception with neuromorphic computation and closed-loop control (see Table~\ref{tab:toy_examples}, green and bold examples)~\cite{Romero_hockey,chen2020event,ziegler_tennis,Rizzo_2025}, the proposed pipeline remains competitive in overall reaction time, while being the fastest to jointly estimate the full state vector, position, speed, and direction.

%% Placing the toy-problem overview table here temporarily
\begin{table}[H]
    \centering
        \caption{Event-based and neuromorphic game applications. Closed-loop neuromorphic perception to action systems in bold, including event-based camera perception input, in bold and green.}
        \label{tab:toy_examples}

        \begin{adjustbox}{max width=1\textwidth}
            \begin{tabular}{*7l}
                    \toprule
                     \textbf{Task} &  \textbf{Reference}  & \textbf{Event-based camera}  & \textbf{SNN} & \textbf{Hardware}  & \textbf{Latency} & \textbf{Power consumption}\\
                    \midrule
                    Robot goalkeeper & Delbruck et al.~\cite{robotgoalie_delbruck} (2013)
                    & DVS128 Tmpdiff128 & no & PC & 3\,ms (sensor-to-actuator reaction time)& 4\% CPU load \\

                    \textcolor{PineGreen}{\textbf{Robot goalkeeper}} & Cheng et al.~\cite{robotgoalie_cheng} (2020)
                    & DVS128 & yes & SpiNN-3 & \textbf{6.5~ms} & 2~mW \\

                     Ball catcher& Wang et al.~\cite{wang_catch} (2022)
                    & Inivation DVXplorer  & no & Nvidia Jetson NX & $\sim$20 ms (network inference) & --  \\

                    \textcolor{PineGreen}{\textbf{Table tennis robot}} &  Ziegler et al.~\cite{ziegler_tennis} (2024)
                    & Prophesee EVK4  & yes & Akida & \textbf{2.20 ± 0.35 ms} (single forward pass) & $\sim$4.5 mW\\
                    & & &                   & DynapCNN & 46.04 ± 1.38 ms (single forward pass)& $\sim$5 mW\\ 
                    & & &                   & Loihi2 & 1458.57 ± 230.50 ms (single forward pass)& $\sim$100 mW\\

                    \midrule

                    \textbf{Atari Pong} & Blakowski et al.~\cite{pong_dreaming_blakowski} (2024)
                    & no camera & yes & DYNAP-SE & -- & -- \\
                    
                    \textcolor{PineGreen}{\textbf{Atari Pong}} & Rizzo et al.~\cite{Rizzo_2025} (2025)
                    & DAVIS346 & yes & RaspberryPi Pico & 11.4 ms & 700 mW \\
                    
                    \midrule
                    Air hockey & Gava et al.~\cite{Gava_PUCK} (2022)
                    & ATIS HVGA Gen3 & no & CPU Intel i7&  \makecell[l]{\textbf{8}~µs \\ (backlog of unprocessed events waiting in the \\ queue, on CPU)} & --  \\

                    \textcolor{PineGreen}{\textbf{Air hockey}} & Romero et al.~\cite{Romero_hockey} (2025)
                    &  Prophesee Metavision EVK3-HD & yes & SpiNN-5 &  \textbf{10.09 ± 1.02~ms} & --  \\

                    \textbf{Air hockey} & Ambrosini et al.~\cite{air_hockey_ambrosini} (2026)
                    & no camera & yes & DYNAP-SE & -- & --  \\

                    \midrule
                    Pencil balancing & Conradt et al.~\cite{Conradt_pencil} (2009)
                    & DVS  & no & CPU & $\sim$50 ms (closed-loop delay) & --  \\
                    \midrule \midrule
                    \textcolor{PineGreen}{\textbf{Pinball}} & \textbf{This work}
                    & Prophesee EVK4  & yes& SpiNN-5& \textbf{5~ms} (I/O \acs{sP} network) 21.7~ms (closed-loop reaction)& \textbf{148 $\mu$W} dyn. power \\
                 \bottomrule
            \end{tabular}
        \end{adjustbox}
\end{table}

% Sim-to-real
The quantitative results were obtained in simulation, where the DVS observed an interactive pinball playfield rendered on a monitor, allowing trajectories and ground truth to be controlled and repeated across runs. Within this controlled setting, the \acs{nPA} loop matched or exceeded human-level hit rates across both flipper regimes while operating within a real-time, low-power budget. To confirm that this behaviour carries over beyond the simulated setting, the loop was additionally deployed on a hardware demonstrator (Section~\ref{sec:physicalPinball}): a metal ball rolling in a box under ambient lighting, observed by the identical DVS, decoded by the same spiking network, and used to drive servo-actuated flippers in closed loop. This confirms that the pipeline operates on a real scene rather than a rendered one, though as a qualitative demonstration rather than a quantitative one; a full quantitative evaluation on physical hardware remains for future work.

The proposed work points to several natural extensions. The perception pipeline is independent of the striking policy choice, kept minimal to evaluate the perception to action loop, and more capable controllers could be paired with it. In particular, learned policies trained on the decoded ball state would be expected to improve gameplay substantially. Similarly, the current single-scale receptive field is tuned to a specific trade-off between slow- and fast-motion sensitivity (Section~\ref{ch:syst_charact}); drawing on the multi-scale organisation of biological visual systems, which combine smaller receptive fields for fine spatial precision with larger ones for fast, coarse motion~\cite{hubel1962receptive, ringach2002spatial}, a comparable multi-scale arrangement would let the model perceive a wider range of motion speeds while improving the precision of motion and velocity estimation. Finally, on the acquisition side, the $60\,\mathrm{Hz}$ display used in this study imposes a temporal-sampling ceiling on the visual stimulus itself, independent of the network; a higher-refresh-rate display would relax this constraint in future evaluations. 

In conclusion, these results indicate that a spiking motion-estimation front end coupled to a lightweight decision policy can sustain closed-loop control under the joint latency and energy constraints that challenge frame-based approaches. Here, competitive performance was achieved with fewer than $25\,$k neurons and sub-milliwatt dynamic power. The presented modular and pipelined architecture provides a blueprint for future neuromorphic systems that must operate under strict power and timing constraints, such as autonomous robotics, human–machine interaction, and adaptive sensorimotor control.

\section{Conclusion}\label{ch:conclusion}

This work presents a fully characterised, spiking, real-time pipeline for closed-loop neuromorphic perception-to-action pinball gameplay, deployed end-to-end on the SpiNNaker neuromorphic platform. From raw DVS events, the spiking-based pinball (\acs{sP}) network extracts the ball's position, speed, and direction entirely through structured spiking computation, with no learned or trained parameters. This state estimate is then converted into flipper commands by a strike decision-making policy. 

The network is systematically characterised, and neuromorphic-perception-to-action (\acs{nPA}) loop performance benchmarked directly against human players across two flipper regimes of increasing physical realism, positioned against prior event-based and neuromorphic closed-loop demonstrators, and validated beyond simulation on a physical demonstrator. Characterisation across receptive field size, accumulation window, and angular tuning width identified $3\times3$ receptive fields, $\Delta T = 16.7\,\mathrm{ms}$, and $\sigma = 30$ as the best configuration for real-time operation. 
In the binary-gate regime, the loop reached a hit rate of $56.1\%$, almost twice the average human performance and close to the best-performing player, while maintaining a low internal network latency of approximately $5\,\mathrm{ms}$, an overall reaction latency of $21.7\,\mathrm{ms}$, and sub-milliwatt dynamic power. This performance was achieved with fewer than $25\,$k neurons and an estimated dynamic power of $\approx 148\,\mu\mathrm{W}$, underlining the efficiency of the fully spiking approach.
Whilst in the solid-flipper regime, a single policy parameter drove the \acs{nPA} loop across the full range of human strategies, from precise to aggressive, with the perception pipeline held fixed. This indicates that the spiking state estimate is accurate enough to support a whole family of viable strike policies, and that closed-loop performance is principally bounded by perception quality rather than policy sophistication. The resulting latency and power figures are further shown to be competitive with, and in several cases superior to, prior event-based and neuromorphic closed-loop demonstrators.

A physical demonstrator, in which a real metal ball is tracked and physical flippers are actuated in a closed loop, confirmed that the same perception-to-action principle operates beyond simulation. Together, these findings show that fully event-driven, spiking-based computation can match human-level responsiveness in a fast visuomotor task while meeting the stringent latency and energy budgets of real-time operation.

Several directions remain open for future work. A more compact network, together with alternative kernel sizes and stride configurations, could further reduce the computational load and power demands of the pipeline. 
In future work, the system can be extended to embodied robotics by integrating the pipeline with the iCub humanoid platform, enabling real-time sensorimotor control in more complex, three-dimensional interactive environments.

\section{Data and Code Availability}\label{ch:code}

The code related to the \acs{sP} network and the PinballSim can be found here: \\
\href{https://gitlab.univ-lille.fr/bioinsp/NeuromorphicPinball}{https://gitlab.univ-lille.fr/bioinsp/NeuromorphicPinball}.\\
The data can be found here: \\\href{https://nextcloud.univ-lille.fr/index.php/s/oTGy5kS5ep73ZN8}{https://nextcloud.univ-lille.fr/index.php/s/oTGy5kS5ep73ZN8} \\
The video can be found at: \href{https://youtu.be/2iy6tDeiyB8}{https://youtu.be/2iy6tDeiyB8}

\section{Acknowledgement}\label{ch:acknowledgement}
This work is supported by IRCICA (Univ. Lille, CNRS, USR 3380 -- IRCICA, F-59000 Lille, France) and by a public grant overseen by the French National Research Agency (ANR) as part of the ``PEPR IA France 2030'' programme (Emergences project ANR-23-PEIA-0002). G.D. and S.P. were additionally supported by the Czech Science Foundation (GAČR) project PIONEER, no. 26-23432S, and S.P. by the ROBOPROX project (reg. no. CZ.02.01.01/00/22 008/0004590).
\printbibliography

\end{document}